\documentclass[10pt,twocolumn,letterpaper]{article}

\usepackage[pagenumbers]{iccv} % To force page numbers, e.g. for an arXiv version

\definecolor{iccvblue}{rgb}{0.21,0.49,0.74}
\usepackage[pagebackref,breaklinks,colorlinks,allcolors=iccvblue]{hyperref}

\usepackage{multirow}
\usepackage{algorithm}
\usepackage{algorithmicx}
\usepackage{algpseudocode}
\usepackage{amsmath}

\newcommand{\algname}{RadarSplat\xspace}
\def\paperID{12641} % *** Enter the Paper ID here
\def\confName{ICCV}
\def\confYear{2025}

\title{RadarSplat: Radar Gaussian Splatting for High-Fidelity Data Synthesis and 3D Reconstruction of Autonomous Driving Scenes}

\author{
Pou-Chun Kung \quad Skanda Harisha \quad Ram Vasudevan \quad Aline Eid \quad Katherine A. Skinner \\
University of Michigan \\
{\tt\small \{pckung, skandah, ramv, alineeid, kskin\}@umich.edu}
}
\begin{document}
\maketitle

\begin{abstract}
High-Fidelity 3D scene reconstruction plays a crucial role in autonomous driving by enabling novel data generation from existing datasets. 
This allows simulating safety-critical scenarios and augmenting training datasets without incurring further data collection costs.
While recent advances in radiance fields have demonstrated promising results in 3D reconstruction and sensor data synthesis using cameras and LiDAR, their potential for radar remains largely unexplored.
%3D reconstruction and sensor data synthesis for radar remain largely unexplored.
Radar is crucial for autonomous driving due to its robustness in adverse weather conditions like rain, fog, and snow, where optical sensors often struggle.
Although the state-of-the-art radar-based neural representation shows promise for 3D driving scene reconstruction, it performs poorly in scenarios with significant radar noise, including receiver saturation and multipath reflection. 
%, due to inadequate noise modeling.
Moreover, it is limited to synthesizing preprocessed, noise-excluded radar images, failing to address realistic radar data synthesis.
To address these limitations, this paper proposes \algname, which integrates Gaussian Splatting with novel radar noise modeling to enable realistic radar data synthesis and enhanced 3D reconstruction.
Compared to the state-of-the-art, \algname achieves superior radar image synthesis ($+3.4$ PSNR / $2.6\times$ SSIM) and improved geometric reconstruction ($-40\%$ RMSE / $1.5\times$ Accuracy), demonstrating its effectiveness in generating high-fidelity radar data and scene reconstruction. A project page is available at \href{https://umautobots.github.io/radarsplat}{https://umautobots.github.io/radarsplat}.
%Our code will be released after the acceptance.
\end{abstract}

\section{Introduction}

Data-driven, learning-based methods have significantly advanced autonomous driving; however, acquiring suitable training data remains a substantial challenge. 
Real-world data collection to train models is time-consuming and prohibitively expensive, while developing realistic sensor simulations during real-world driving scenarios is hindered by the persistent simulation-to-reality gap.

Recent advances in 3D scene reconstruction using neural radiance field (NeRF) and Gaussian Splatting (GS) enabled closed-loop simulation~\cite{unisim} and synthetic data generation~\cite{neurad, Autosplat, lihigs}, which are critical for enhancing data-driven autonomous driving systems. While radar is an essential sensor for modern autonomous vehicles, most existing 3D scene reconstruction and data synthesis methods focus exclusively on cameras and LiDAR~\cite{NSG, neurad, drivinggs}, leaving radar’s potential largely unexplored. %Radar stands out for its low cost, ability to measure velocity, and robustness in challenging weather conditions, where optical sensors usually fail. 
Radar can enable weather-resilient SLAM~\cite{allweatherradarslam, allweatherradarslam2, radar_survey}, localization~\cite{allweatherradarslam3, radar_reloc, ndt_ro} and perception~\cite{kradar, radiate, radarocc}, and can boost performance for multimodal perception tasks~\cite{cramnet, simplebev, crkd, lirafusion, bilrfusion}.

%Radar stands out for its low cost, ability to measure velocity, and robustness in challenging weather conditions such as rain, fog, and snow, where optical sensors struggle.
%These capabilities have played a key role in the adoption of radar sensors across myriad research areas in autonomous driving and robotics, including weather-resilient SLAM~\cite{allweatherradarslam, allweatherradarslam2}, odometry~\cite{allweatherradarslam3, ndt_ro}, and re-localization~\cite{radar_reloc}. 
%However, these approaches are limited to 2D scene reconstruction and pose estimation, reducing their effectiveness in 3D environments.
%Research on radar-based object detection has also been conducted~\cite{kradar, radiate}. 
%Moreover, radar has proven beneficial when fused with cameras and LiDAR for bird's-eye-view (BEV) %segmentation~\cite{simplebev} and object detection~\cite{crkd, lirafusion, bilrfusion}. 
%Still, generating realistic radar data for training remains a challenge, limiting the performance of learning-based radar models.

Recently, DART~\cite{dart} achieved realistic data synthesis of range-Doppler radar signals using a NeRF. 
However, dense range-Doppler images are uncommon in radar systems for autonomous driving.
In contrast, Radar Fields~\cite{rf} designed a NeRF-based approach for 3D reconstruction and novel view synthesis using scanning radar images, which are widely used in autonomous driving research~\cite{oxford_robotcar, boreas, radiate, mulrun, hercules}. 
While Radar Fields demonstrates encouraging results, due to the lack of noise modeling, it can only synthesize preprocessed, noise-excluded radar images, making realistic radar data synthesis still a challenging problem.

This work proposes a GS-based framework for radar with noise detection and modeling. 
The proposed radar rendering formulation integrates azimuth and elevation antenna gain and accounts for spectral leakage to model the blurred radar range measurement. 
The radar noises are handled by view-dependent noise probability and periodic modeling.
This enables more realistic radar image synthesis and improved 3D geometry estimation compared to ~\cite{rf}, as shown in Figure~\ref{fig:teaser}.
The key contributions of this paper are:

\begin{itemize}
    \item The first 3D Gaussian Splatting formulation for radar in autonomous driving.
    \item High-fidelity radar data synthesis that incorporates modeling of multipath effects and noises.
    \item Efficient radar noise detection and a denoising method that improves signal clarity and occupancy estimation.
    \item Radar inverse rendering that disentangles the real target, noise, and multipath effects from noisy radar images.
\end{itemize}

\begin{figure*}[t!]
    \centering
    \includegraphics[width=0.95\linewidth]{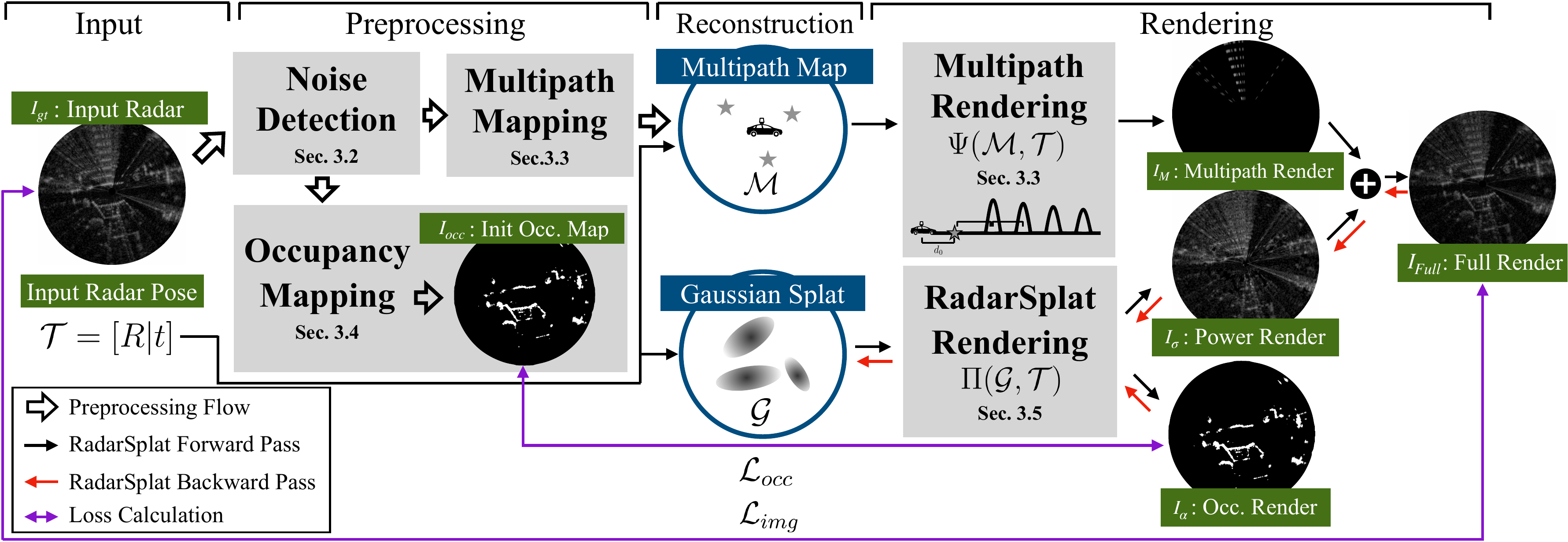}
        \vspace{-0.05in}
        \caption{System Overview. \algname takes radar images and poses as input. The preprocessing step includes noise detection and initial occupancy mapping. The multipath source map and Gaussian splat reconstruct the 3D scene and model multipath effects for novel view synthesis.}
        \vspace{-0.05in}
        \label{fig:system_diagram}
\end{figure*}

\section{Related Works}
\subsection{Radiance Fields for Data Synthesis}

Radiance field methods have opened a new era of high-fidelity reconstruction and novel view synthesis using cameras. NeRF \cite{nerf} first introduced an implicit neural scene representation and image rendering approach. This work was soon extended to autonomous driving scenes for data synthesis and closed-loop simulation~\cite{NSG, MARS, emernerf, unisim}.

More recently, Gaussian Splatting (GS)~\cite{3dgs} replaced the implicit neural network with explicit 3D Gaussians for scene representation, offering faster and more accurate reconstruction and data synthesis. GS has also been adapted for autonomous driving applications~\cite{pvg, streetgs, drivinggs}, facilitating the generation of high-quality autonomous driving data.
% , drivinggs, autosplat, HUGS
\subsection{LiDAR-Integrated Radiance Fields}
While cameras are the primary sensors used by radiance field methods, other sensors have also been incorporated to improve the synthesis of sensor data. Several studies have explored radiance fields with LiDAR/depth measurement~\cite{lidar-nerf, NFL, dynNFL, lidar4d, loner, sadgs, depthgs}.
Recent research on LiDAR-camera integration~\cite{urf, cloner, neurad} has shown that using LiDAR measurements helps improve 3D scene reconstruction and leads to better image rendering.
LiDAR-integrated GS methods have also been introduced~\cite{lidargs, ligs, tclcgs, liv-gaussmap, letsgo, lihigs}, combining the strengths of GS for more accurate scene reconstruction and data synthesis.

\subsection{Radar and Sonar-Integrated Data Synthesis}
In addition to cameras and LiDAR, researchers have explored integrating non-optical sensors such as radar and sonar into NeRF/GS. Unlike optical sensors that provide images or point clouds, radar and sonar produce range-power images, requiring redesigned sensor models for rendering.
%radiance field rendering.
Sonar NeRF/GS methods~\cite{sonar_nerf, sas_nerf, sonarsplat} have been developed for 3D reconstruction in underwater environments. More recently, Z-Splat~\cite{zsplat} introduced a camera-sonar fusion GS approach to enhance indoor scene reconstruction.

On the other hand, most radar-based NeRF methods~\cite{sar_nerf, isar_nerf, ranerf, sar_rf} focus on 3D reconstruction using synthetic aperture radar (SAR). These methods require a specific sensor trajectory (usually linear or circle) to simulate a much larger antenna for high-resolution imaging. This limits their applicability to satellites or aircraft, making them unsuitable for automotive use. Recently,~\cite{weather_radar} proposed weather radar forecasting with a Gaussian representation, but their method approximates the radar model through camera image projection, and is not designed for scene reconstruction and data synthesis. 
DART~\cite{dart} introduced a NeRF-based method using a TI MIMO radar, showing promising results for scene reconstruction and novel view synthesis in indoor, room-scale environments. However, it requires a range-Doppler image, which is not available from radars commonly used in autonomous driving.

Radar Fields~\cite{rf}, the work closest to ours, proposes a NeRF-based method for scanning radar, achieving promising scene reconstruction and radar image rendering. 
However, it does not model the impact of radar noise, making the method unreliable in high-noise conditions, which are common in real-world autonomous driving. Additionally, its training data is denoised using a dynamic threshold, preventing the model from accurately simulating noise in synthetic radar images. This limitation reduces the accuracy of radar data synthesis. In contrast, we introduce noise decomposition and modeling to improve occupancy estimation and enable noise synthesis from novel views.

%More recently, ~\cite{weather_radar} proposed modeling weather radar with Gaussian representation as a compact input for a weather forecasting network. However, their method approximates radar model through camera image projection without considering radar sensing modality, which is not design for scene reconstruction and data synthesis.

\subsection{Radar Data Simulation and Synthesis}
%There has long been interest in simulating radar measurements. The most accurate method for simulation is to use electromagnetic full-wave solvers, which precisely model the interaction of radar signals with various materials and shapes. However, this approach relies on iteratively solving Maxwell's equations to converge to an accurate solution, making it highly time-consuming and computationally intensive. So in order to save time and resources researchers have proposed model-based approaches to simulate radar signal propagation using physics-based environmental models, often leveraging ray tracing techniques~\cite{ray_tracing_radar, ray_tracing_radar2, sar_radar_sim}.
Simulating radar measurements has long been of interest. The most accurate method uses electromagnetic full-wave solvers, but iteratively solving Maxwell's equations is computationally expensive. To save time and resources, researchers have developed model-based simulations using physics-driven environmental models and ray tracing~\cite{ray_tracing_radar, ray_tracing_radar2, sar_radar_sim}.
While these simulators capture complex radar phenomena such as occlusion, path loss, and multipath effects, they rely on manually defined scenes, limiting realism. Additionally, data-driven methods reconstruct real-world scenes from sensor data~\cite{yaw_rio}, but typically produce sparse radar points based on constant false alarm rate (CFAR) detection~\cite{cfar}. In contrast, the proposed method synthesizes dense, high-fidelity radar images without information loss.

\section{Methods}
This section first introduces the radar sensing equation and common noise types in radar images (Sec.~\ref{sec:radar_sensing}). 
Figure~\ref{fig:system_diagram} provides an overview of our pipeline. To account for radar noise, we propose a noise detection method (Sec.~\ref{sec:multipath_saturation_detection}) and model multipath effects by identifying their sources, enabling recovery in novel view rendering (Sec.~\ref{sec:multipath_modeling}).
Based on noise detection, we develop a radar denoising approach for noise-free occupancy mapping, serving as a supervisory signal for \algname (Sec.~\ref{sec:occ_mapping}).
For scene reconstruction, we present a radar model that renders radar images from 3D Gaussians based on radar physics (Sec.~\ref{sec:radargs}). 
Finally, we define the \algname training loss (Sec.~\ref{sec:loss}).

\subsection{Radar Sensing Foundation}
\label{sec:radar_sensing}
\subsubsection{Radar Equation}
A radar image is comprised of multiple range-power signals captured from different azimuth angles.
In a range-power signal, the received power $P_r(n)$ at bin $n$ with range $R_n$ is determined by the radar equation~\cite{radar_equation}:
\begin{equation}
    P_r(n) = \frac{P_t G^2  \lambda^2 \sigma }{(4\pi)^3 R_n^4 L},
\label{eq:radar_equation}
\end{equation}
where $P_t$ is the radar peak transmit power, $G$ is the antenna gain, $\lambda$ is the wavelength, $L$ represents system and propagation loss, and $\sigma$ is the total radar cross-section of objects within the radar wave at range $R_n$.

\begin{figure}[t!]
    \centering
    \includegraphics[width=0.99\linewidth]{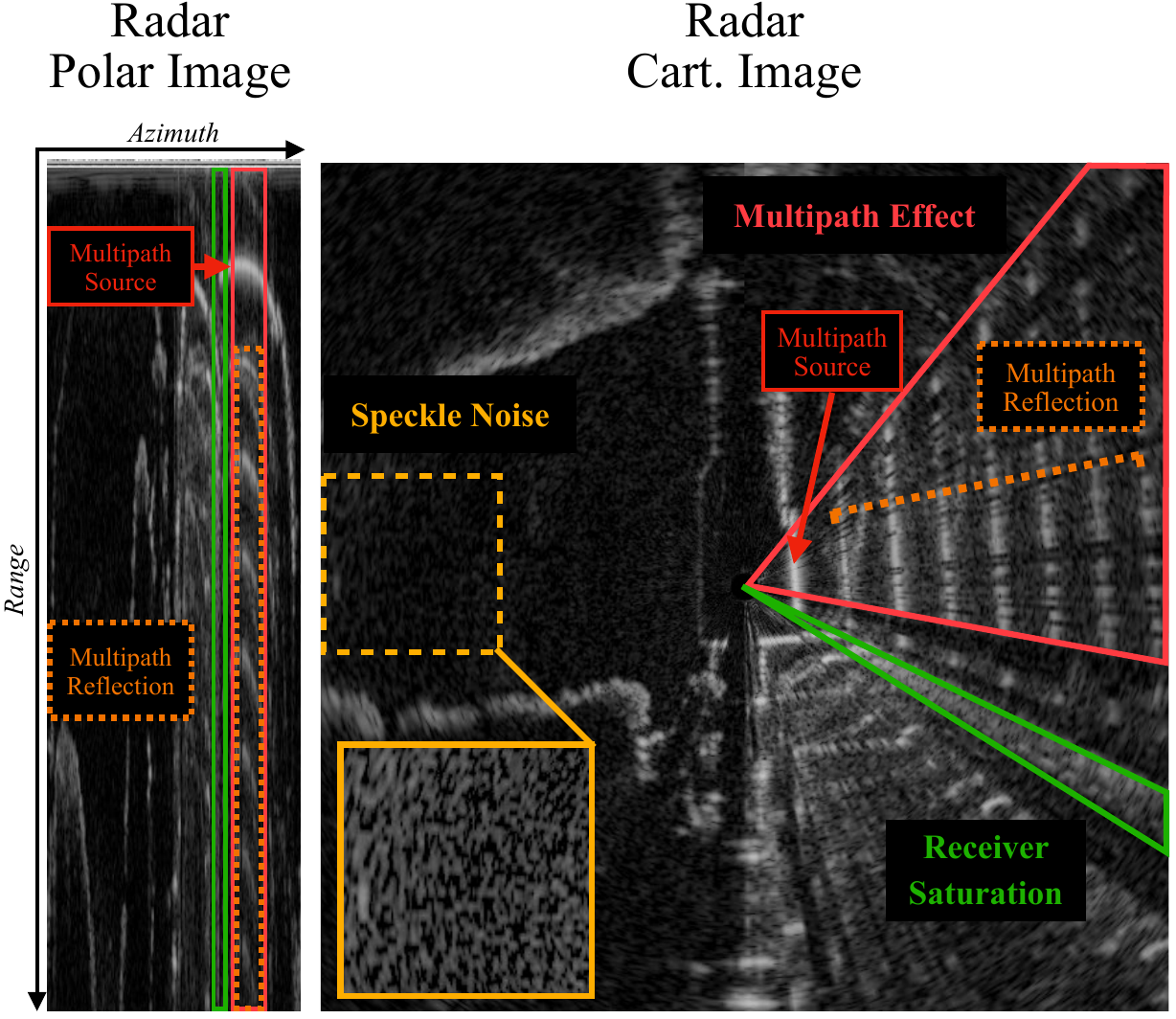}
        \vspace{-0.1in}
        \caption{Three types of radar noise of scanning radar highlighted in a raw radar image in polar space (bottom) and Cartesian space (top). The source and ghost reflections of multipath effects are highlighted with red and orange boxes. 
        % The polar image is reshaped for better visualization.
        }
        \vspace{-0.1in}
        \label{fig:radar_noises}
\end{figure}

\subsubsection{Radar Noise Types}
%Radar systems are affected by various types of noise that degrade performance and accuracy. 
Figure ~\ref{fig:radar_noises} illustrates three common radar noises. 
\textit{Multipath effects} (red)
occur when radar signals reflect off multiple surfaces before reaching the receiver, causing multiple echoes to arrive at different times. This results in periodic ghost targets appearing behind real targets.
\textit{Receiver saturation} (green)
occurs when a strong signal return overwhelms the radar receiver, causing signal distortion and resulting in a uniform power offset in radar images. 
\textit{Speckle noise} (yellow)
is a granular interference pattern that appears in radar images due to the coherent nature of radar signals, appearing as weak background noise.
\begin{figure}[t!]
    \centering
    \includegraphics[width=0.99\linewidth]{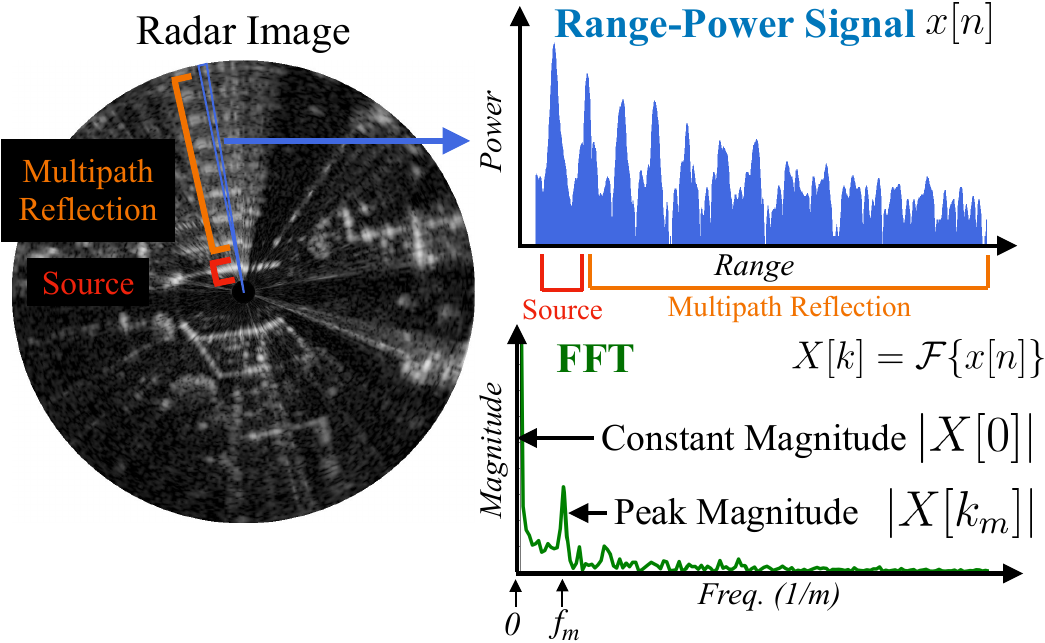}
        \vspace{-0.1in}
        \caption{Range-power signal and its FFT of a radar azimuth beam with multipath effects. The constant and peak magnitude in the FFT results are used for noise detection and denoising.}
        \label{fig:fft}
        \vspace{-0.1in}
\end{figure}

\begin{figure*}[t!]
    \centering
    \includegraphics[width=0.99\linewidth]{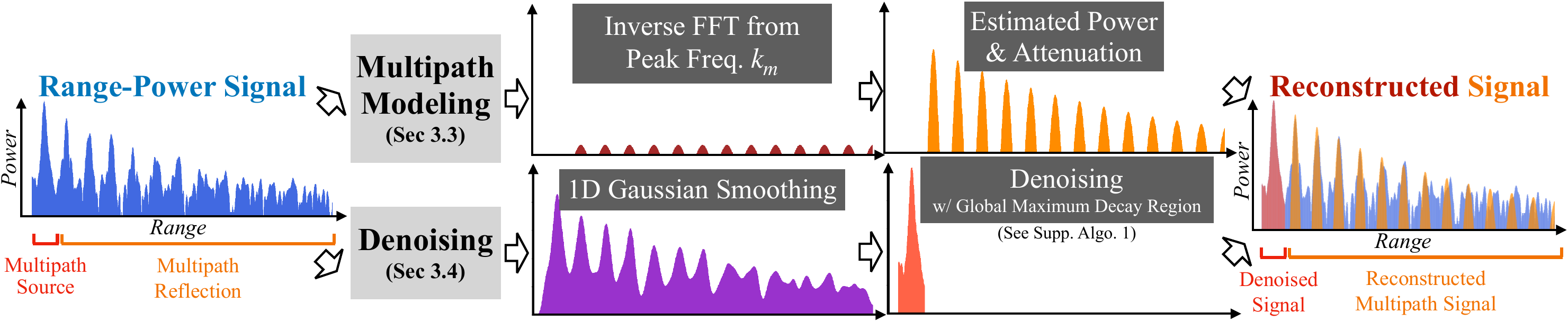}
        \caption{Multipath modeling and denoising. The multipath effect is modeled by peak frequency and source power reflection and attenuation. The denoising method removes ghost detection of saturation and multipath effects, and the denoised image is used to construct an occupancy map to guide \algname training. Note that the denoising is apply to both detected saturated and multipath beams.}
        \label{fig:denoising}
\end{figure*}

\subsection{Multipath and Saturation Noise Detection}
\label{sec:multipath_saturation_detection}
Unlike ~\cite{rf}, which removes noise using a dynamic threshold that does not adapt to the radar equation, we instead detect and model radar noise to synthesize more realistic radar images. We adopt a noise detection method based on the Fast Fourier Transform (FFT), as in Figure~\ref{fig:fft}. First, we apply FFT to all azimuth beams:
% \begin{equation}
%     X[k] = \mathcal{F}\{x[n]\}
% \end{equation}

% Fourier Transform
\begin{equation}
    X[k] = \mathcal{F} \{ x[n] \} = \sum_{n=0}^{N-1} x[n] e^{-j \frac{2\pi}{N} kn},
\end{equation}
% Inverse Fourier Transform
% \begin{equation}
% x[n] = \mathcal{F}^{-1} \{ X[k] \} = \frac{1}{N} \sum_{k=0}^{N-1} X[k] e^{j \frac{2\pi}{N} kn}, \quad n = 0, 1, \dots, N-1
% \end{equation}
where $x[n]$ represents the range-power signal from a row in a radar image, and $n$ is the range bin index with a total of $N$ bins. $X[k]$ is the frequency domain output from the discrete Fourier transform, with $k$ as the frequency index.

\textbf{Receiver Saturation}. A beam experiencing receiver saturation typically exhibits a uniform radar power offset across all ranges. This uniform power distribution results in a constant term in the frequency domain. Based on this observation, we define a constant ratio value, $\mathcal{C}$:
\begin{equation}
    \mathcal{C} = \frac{|X[0]|}{\sum_{k=1}^{N-1} |X[k]|}
\end{equation}
Beams with $\mathcal{C} > \mathcal{C}_{th}$ are classified as saturation beams, $\Theta_{sat}$, where $C_{th}$ is an experimentally determined magnitude threshold.

\textbf{Multipath Effects}.  Multipath noise appears as a fixed periodic pattern in a radar beam. This periodic power pattern contributes to a significant amplitude peak in the frequency domain (Figure~\ref{fig:fft}). Based on this observation, we identify the significant frequency with index $k_m$ that has maximum magnitude.
% \begin{equation}
%     k_m = \arg\max_{k \neq 0}
% \end{equation}
Beams with magnitude $|X[k_m]| > A_{th}$ and a constant value ratio $\mathcal{C} > \mathcal{C'}_{th}$ are classified as multipath beams, $\Theta_{multi}$. %The magnitude threshold $A_{th}$ is a magnitude threshold is set experimentally. 
$A_{th} \text{ and }C'_{th}$ are experimentally determined magnitude thresholds. Note that since multipath typically occurs alongside saturation beams, we use a relaxed criteria $\mathcal{C'}_{th}$ to reduce false positive multipath detection. The azimuth angles of detected noisy beams, $\Theta_{sat}$ and $\Theta_{multi}$, are saved for later use in multipath modeling and occupancy mapping.

\subsection{Modeling Multipath Effects}
\label{sec:multipath_modeling}

After detecting the multipath beams, we aim to model the noise source by estimating its position, reflectivity, and power attenuation, allowing us to simulate multipath effects from novel views. Figure~\ref{fig:denoising}  illustrates our proposed pipeline for noise source modeling. First, the source distance $d_m$ is determined from the peak frequency index, $k_m$:
\begin{equation}
    f_m = \frac{k_m}{N \Delta r} \quad \text{(1/m)}
\label{eq:k2freq}
\end{equation}
\begin{equation}
    d_m = 1/f_m \quad \text{(m)}
\label{eq:freq2dist}
\end{equation}
where $\Delta r$ is range resolution.
With known azimuth angle $\theta_m \in \Theta_{multi}$, distance $d_m$, and radar pose $\mathcal{T}$, we can get the multipath source location, $\mu$, activated view angle, $\theta_{view}$, and range, $r_{view}$. Therefore, a rough multipath signal reconstruction can be obtained by performing an inverse Fourier Transform with magnitude $|X[k_m]|$ and phase $\angle X[k_m]$:
% half-spectrum inverse Fourier Transform

\begin{equation}
    x_{\text{m}}[n] = \frac{1}{N} |X[k_m]| \cos\left( \frac{2\pi}{N} k_m n + \angle X[k_m] \right)
\label{eq:inverse_fft}
\end{equation}
To further reconstruct multipath effects from novel views, we then define the power reflection, $A_m$, and exponential attenuation rate, $\gamma_m$, of the source, so that the multipath effect is represented as:
\begin{equation}
    x'_m[n] = A_m e^{-\gamma_m n} x_m[n]
\label{eq:A_gamma}
\end{equation}
$A_m$ and $\gamma_m$ can be estimated by fitting raw data, $x[n]$, with least squares.
% \begin{equation}
%     \min_{A_m, \gamma_m} \sum_{n=0}^{N-1} (x[n] - x'_m[n])^2
% \end{equation}
Finally, the map of multipath sources is stored as:
\begin{equation}
    \mathbf{\mathcal{M}} = \{\mathcal{M}_i:( \mu^i, \theta_{view}^i, r_{view}^i, A_m^i, \gamma_m^i)\ | i=1,...,l \}.
\end{equation}

When rendering from novel viewpoints, $\mathcal{T}_{novel}$, we compare the new view angle and distance differences, $\Delta r_{view}$ and $\Delta \mathcal{\theta}_{view}$, to determine if multipath effects are present. If $\Delta r_{view} < r_{th}$ and $\Delta \mathcal{\theta}_{view} < \theta_{th}$, we reconstruct multipath effects $x'_m[n]$ from the novel view using the updated $d_m$ and Eq.~\ref{eq:k2freq}-\ref{eq:A_gamma}, % , eq:freq2dist, eq:k2freq, eq:inverse_fft, eq:A_gamma
 where $r_{th}\text{ and }\theta_{th}$ are experimental thresholds. 
The multipath rendering is denoted as $\Psi$, where the multipath image from a given view $\mathcal{T}$ is expressed as $I_{\mathcal{M}} = \Psi(\mathcal{M}, \mathcal{T})$.

% \begin{figure}[t!]
%     \centering
%     \includegraphics[width=0.99\linewidth]{pdf/signal_processing_v4.pdf}
%         \caption{Multipath modeling and noise removal. The multipath effect is modeled by peak frequency and source amplitude and attenuation. The denoising method removes ghost detection, and the denoised image is used to construct an occupancy map to guide \algname training.}
%         \label{fig:denoising}
% \end{figure}

\subsection{Denoising and Occupancy Map Pre-processing}
\label{sec:occ_mapping}

\begin{figure}[t!]
    \centering
    \includegraphics[width=0.99\linewidth]{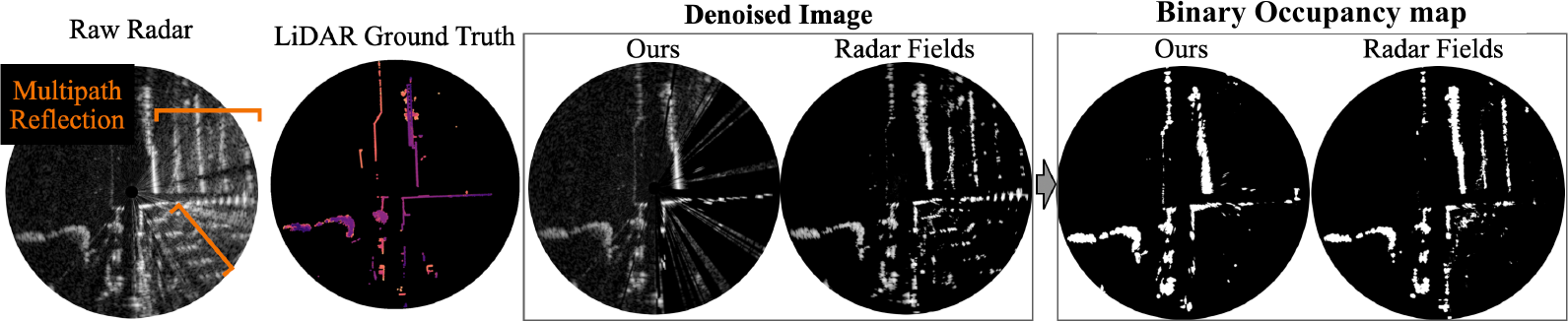}
        \caption{Our proposed radar image denoising method preserves rich information while remaining robust to multipath effects. In contrast, the dynamic threshold approach used in~\cite{rf} struggles with multipath scenarios that include strong power returns. The remaining low-power speckle noise in the denoised image is later removed by binary occupancy mapping.}
        \label{fig:denoise_results}
\end{figure}

Following Radar Fields~\cite{rf}, we use an initial estimated occupancy map to guide training, which helps to disentangle noise and real objects in radar images. %But instead of relying solely on a dynamic threshold, our noise detection (Sec.~\ref{sec:multipath_saturation_detection}) enable a more robust noise removal, leading to improved occupancy maps for training. 
We propose a denoising algorithm that removes noise across detected noisy azimuth angles, $\theta_{noise} \in \Theta_{sat} \cup \Theta_{multi}$, identified in Sec.~\ref{sec:multipath_saturation_detection}.  The process involves Gaussian smoothing of the raw signal, identifying the distance with the maximum magnitude, and searching the decay region to generate a noise-free mask. The denoising process is illustrated in Figure~\ref{fig:denoising}, with the pseudo-code provided in the supplementary material.
% $G_{\sigma}(n) = \frac{1}{\sqrt{2\pi\sigma^2}} e^{-\frac{n^2}{2\sigma^2}}$
% $x_{smooth} = (x*G_{\sigma})$
Figure~\ref{fig:denoise_results} shows qualitative results of our proposed denoising method. Our method produces a clear denoised image, whereas Radar Fields struggles with multipath effects. The denoised image is then used to generate an initial occupancy map following ~\cite{rf}. The denoised image and occupancy map are shown in Figure~\ref{fig:denoise_results}.

\subsection{Radar Gaussian Splatting}
\label{sec:radargs}

% \begin{figure}[t!]
%     \centering
%     \includegraphics[width=0.99\linewidth]{pdf/radar_rendering_v3.pdf}
%         \caption{\algname rendering. We first project Gaussians to 2D using the elevation antenna gain, then apply the azimuth antenna gain via a 1D convolution along the azimuth axis. Radar spectral leakage is modeled using 1D Gaussian smoothing along the range axis.}
%         \label{fig:radar_rendering}
% \end{figure}

We represent the scene as a set of 3D Gaussians following~\cite{3dgs}. The model $\mathcal{G}$ has $N$ Gaussians, and each Gaussian is composed of the mean, $\mu$, rotation quaternion, $q$, scaling vector, $S$, and radar power return ratio, $\sigma$, which is the radar cross-section (RCS) value when a Gaussian represents a real object rather than noise: 
\begin{equation}
    \mathcal{G} = \{G_i:( \mu_i, q_i, S_i, \sigma_i)\ | i=1,...,N \}.
\end{equation}

Prior work~\cite{rf} strictly defines the power return ratio as the RCS value, $\sigma_i=\alpha_i \cdot \rho_i$, which fails to account for radar noise, where $\alpha_i$ represents occupancy probability and $\rho_i$ represents reflectance. Instead, we compute the power return ratio, $\sigma_i$, by considering an additional noise probability term, $\eta_i$:
\begin{equation}
    \sigma_i = \rho_i \cdot \min(\alpha_i + \eta_i, 1)
\label{eq:power_return_ratio}
\end{equation}
To account for view dependency, the reflectivity is defined by spherical harmonics (SH) dependent on the view angle: $\rho_i=SH(\theta_{view})$. Additionally, a regularization loss term is designed to make $\alpha_i$ and $\eta_i$ sum to one.

Next, we introduce our rendering pipeline, which incorporates elevation and azimuth projection along with spectral leakage modeling. An overview illustration is provided in Figure 6 of the supplementary material.
%Figure~\ref{fig:radar_rendering} illustrates our radar image rendering pipeline, which applies elevation and azimuth projection along with radar uncertainty modeling.

% \begin{figure}[t!]
%     \centering
%     \includegraphics[width=0.9\linewidth]{pdf/antenna_gain.pdf}
%         \caption{Elevation and Azimuth Antenna Gain of Navtech scanning radar.}
%         \label{fig:antenna_gain}
% \end{figure}

\subsubsection{Rendering with Elevation Projection} 
To render radar images in polar space, we first convert all Gaussians into spherical coordinates $(r_i, \theta_i, \phi_i) = F(x_i, y_i, z_i)$, following ~\cite{lihigs}. % with:
% Move to Supp.
% \begin{align}
% \mu_{spherical}=
% \begin{bmatrix}
%     r \\
%     \theta \\
%     \phi
% \end{bmatrix}
% =
% \begin{bmatrix}
%     \sqrt{x^2 + y^2 + z^2} \\
%     \arctan2(y, x) \\
%     \arcsin\left(\frac{z}{r}\right)
% \end{bmatrix}
% \end{align}
% \begin{equation}
%     \Sigma_{\text{spherical}} = J \Sigma J^T
% \end{equation}
% where $J$ is the Jacobian of Cartesian-to-spherical space conversion.
To obtain the radar image, a 3D-to-2D Gaussian projection is applied to the $\phi$-axis. The rasterization step used in~\cite{3dgs} is simplified by accumulating the weighted power return ratio, $\sigma$, for all Gaussians having the same $r$ and $\theta$. $\sigma$ is weighted by the radar elevation antenna gain profile, $G_{\phi}$ %(Figure~\ref{fig:antenna_gain})
. The rendering equation is defined as:
\begin{equation}
    P_r(\theta, n) = \sum_{\phi_i}^{i}\frac{P_t \cdot G(\phi_i)^2 \cdot \sigma_i}{(4\pi)^3 R_n^4}
\label{eq:elev_render}
\end{equation}
Note that the wavelength, $\lambda$, and the general loss factor, $L$, in Eq.~\ref{eq:radar_equation} are ignored here since they are both constants.
%, and we are not estimating absolute RCS value in this work. 
After the projection, we obtain an intermediate rendering, $I_{Elev}$. 
%To render occupancy or noise probability, $\sigma_i$ is be replaced by $\alpha_i$ or $\eta_i$ in Eq.~\ref{eq:elev_render}.
\subsubsection{Rendering with Azimuth Projection}
Radar typically has a wide beamwidth; in our experiments, the scanning radar covers $1.8^{\circ}$ in azimuth with a $0.9^{\circ}$ scanning resolution. To address this, we define a scaling factor, $Q$, along the image height ($\theta$-axis). Given a ground truth image of size $H \times W$, we render polar images with an intermediate azimuth resolution, $HQ$, using elevation projection, ensuring $I_{Elev} \in \mathbb{R}^{HQ\times W}$. Azimuth projection is then applied via a 1D convolution along the azimuth axis with kernel size $2Q$ and stride size $Q$, and a kernel weighted by the azimuth antenna profile, $G_{\theta}(\theta)
$%, as shown in Figure~\ref{fig:antenna_gain}
, producing the output image $I_{Azi} \in {\mathbb{R}^{H \times W}}$.

\begin{figure}[t!]
    \centering
    \includegraphics[width=0.99\linewidth]{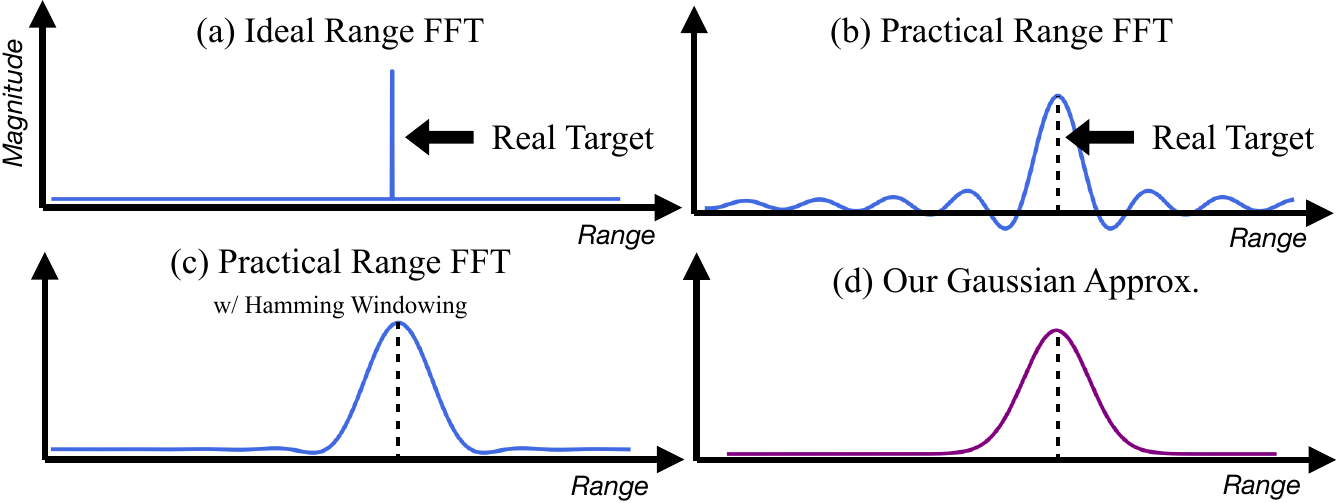}
        \vspace{-0.1in}
        \caption{Modeling spectral leakage in the radar-power signal. (a) Ideal range FFT. (b) Practical range FFT with spectral leakage. (c) Practical range FFT sharpened by a Hamming window. (d) Proposed Gaussian approximation for Hamming-window-sharpened FFT.}
        \vspace{-0.15in}
        \label{fig:spectral_leakage}
\end{figure}

\subsubsection{Spectral Leakage Modeling}
\label{sec:spectral_leakage_modeling}
Radar range measurements are usually blurry due to spectral leakage. %, as shown in Figure~\ref{fig:spectral_leakage_fig}.
%The spectral leakage is due to finite-time sampling of received signals, causing energy to spread across adjacent frequencies when doing range FFT. 
%It makes the object measurement become a Sinc function Figure~\ref{fig:spectral_leakage}-b). 
The spectral leakage is due to finite-time sampling of received signals.
%In practice, radar manufacturers apply a windowing technique to obtain sharp frequency cutoff and lower sidelobes. The radar used in our experiments uses Hamming windowing, which has a range FFT result shown in Figure~\ref{fig:spectral_leakage}-c.
% We observe that the radar signal after windowing is similar to a Gaussian distribution. 
We use a Gaussian distribution to approximate the blurred effect, as shown in Figure~\ref{fig:spectral_leakage}. More details about the spectral leakage modeling are provided in supplemental materials.

%The width of the Sinc function in the frequency domain is $f_w = \frac{2\pi}{T_s}$, where $T_s$ is the sampling duration of the radar.
%It can be converted to the width of a Gaussian in meters with $d = \frac{cf_w}{2\mu}$. We define the variance of the Gaussian as $\sigma_w=d/3$. Therefore the Gaussian approximation is $\sim \mathcal{N}(0, \sigma_w^2)$.

%where $\mathcal{T}=565 uS$ and $\mu=1.6\times10^{12}$. 
%we get $d\approx1.04 (m)$, which is the width of Gaussian

\subsubsection{Final Radar Rendering}
Details
Finally, the complete radar rendering process is denoted as $\Pi$. The radar image is obtained by $\sigma$-rendering, given by $I_{\sigma} = \Pi_\sigma(\mathcal{G}, \mathcal{T})$. The occupancy state is derived via $\alpha$-rendering by replacing $\sigma$ in Eq.~\ref{eq:elev_render} with $\alpha$, expressed as $I_{\alpha} = \Pi_\alpha(\mathcal{G}, \mathcal{T})$. The radar inverse rendering process decomposes the noise-free image $I_{target}$ and the noisy image $I_{noise}$ through $\rho \alpha$-rendering, $\Pi_{\rho \alpha}$, and $\rho \eta$-rendering, $\Pi_{\rho \eta}$.

%To render occupancy or noise probability, $\sigma_i$ is be replaced by $\alpha_i$ or $\eta_i$ in Eq.~\ref{eq:elev_render}.

\subsection{Training Losses}
\label{sec:loss}

The total loss to train the 3D Gaussian scene representation is as follows:
\begin{equation}
\begin{split}
    \mathcal{L} = \lambda_1 \mathcal{L}_{l1} + \lambda_{2} \mathcal{L}_{ssim} +  \lambda_{3} \mathcal{L}_{occ} \\ + \lambda_{4} \mathcal{L}_{size} + \lambda_{5} \mathcal{L}_{reg}
\end{split}
\end{equation}
% The loss function consists of multiple components to ensure accurate rendering and proper constraints on the learned Gaussians.
The $\mathcal{L}_{l1}$ and $\mathcal{L}_{ssim}$ components represent the L1 error and SSIM score between the rendered image $I_{render}$ and ground truth radar image $I_{gt}$. Additionally, $\mathcal{L}_{occ}$ corresponds to the L1 error between the rendered occupancy state $I_{\alpha}$ output by \algname and the initial occupancy map $I_{occ}$ estimated in the preprocessing step to aid in training. $\lambda_i$ are weights for different loss terms.

% To further refine the model, we introduce two regularization loss terms. $\mathcal{L}_{size}$ constrains the maximum size of the Gaussians. This is needed for two reasons: First, the transformation from Cartesian to Spherical Gaussian coordinates relies on a first-order Taylor series approximation, which introduces significant distortion when the Gaussian size is large. Second, during the elevation projection process, the Gaussian center is used to determine the corresponding elevation antenna gain. If the Gaussian is only partially within the radar’s field of view (FOV), this approximation can be inaccurate. By limiting the Gaussian size, we mitigate these issues and ensure a more reliable representation.

To refine the model, we introduce two regularization losses. $\mathcal{L}_{size}$ constrains Gaussians to have a maximum size, $s_{max}$, to prevent distortions from the first-order Taylor approximation in the Cartesian-to-Spherical transformation.
%and inaccuracies in elevation projection when Gaussians partially fall outside the radar's field of view (FOV). 
%This ensures a more reliable representation.
In addition, $\mathcal{L}_{reg}$ is defined as $\mathcal{L}_{reg} = \text{ReLU}(\alpha+\eta-1)$ 
%\begin{equation}
%    \mathcal{L}_{reg} = \text{ReLU}(\alpha+\eta-1)
%\end{equation}
to ensure that the sum of the occupancy and noise probability is smaller than one.

\section{Experiments}
\subsection{Experimental Setup}
We train and test our proposed \algname on the public Boreas Dataset~\cite{boreas}, which includes a Navtech $360^{\circ}$ CIR304-H scanning radar, a 128-beam LiDAR, and ground truth poses from a GNSS/IMU sensor. Following~\cite{rf}, we select every 5 frames as the test frame to create a train-test split.
%withhold $20\%$ of all frames to create a train-test split. 
We selected 13 driving sequences, comprising 7 sunny, 2 night, 2 rainy, and 2 snowy scenes.
% We selected four challenging driving sequences, emphasizing severe radar noise effects caused by surrounding metallic objects such as road signs, light poles, and buildings with concrete and metallic structures, which are common in urban scenarios. 
Each selected sequence has $>10$ seconds duration, which contains more than 40 radar frames. 
% For method configuration, we set $\lambda_1=0.8, \lambda_2=0.2, \lambda_3=5, \lambda_4=10^2, \lambda_5=10^2$. \algname is initialized with $2\times10^4$ Gaussians of size $s=0.5~m$ and trained for $3000$ iterations. For multipath modeling, we set $C_{th}=0.21, A_{th}=0.3, C'_{th}=0.2, r_{th}=0.5m, \text{ and } \theta_{th}=10^\circ$. Gaussian rendering is set to $Q=10$. 
% More details in Supplementary.

% TODO: Move table_scenes to Supp.

\begin{table}[!b]
\centering
\setlength{\tabcolsep}{5pt}
\renewcommand{\arraystretch}{1.3}

\resizebox{1.0\linewidth}{!}{%
\begin{tabular}{lcccclccc}
\hline
\multirow{2}{*}{Method} &  & \multicolumn{3}{c}{Image Synthesis}                 &  & \multicolumn{3}{c}{Scene Reconstruction}                                  \\ \cline{3-5} \cline{7-9} 
                        &  & PSNR$\uparrow$ & SSIM$\uparrow$ & LPIPS$\downarrow$ &  & RMSE$\downarrow$ & R-CD.$\downarrow$ & \multicolumn{1}{l}{Acc.$\uparrow$} \\ \cline{1-1} \cline{3-5} \cline{7-9} 
Radar Fields             &  & 22.66          & 0.20           & 0.60              &           & 3.03             & 0.29              & 0.59                               \\
Ours                    &  & \textbf{26.06} & \textbf{0.51}  & \textbf{0.37}     &           & \textbf{1.81}    & \textbf{0.04}     & \textbf{0.91}                      \\ \hline

\end{tabular}
}
\vspace{-0.1in}
\caption{\footnotesize Image synthesis and geometry reconstruction evaluation on Boreas dataset~\cite{boreas}. Image synthesis is evaluated from unseen views, and geometry reconstruction is evaluated against the LiDAR map. Two snowy scenes are excluded from geometry evaluation due to LiDAR inaccuracy.}
\label{tab:vs_radarfields}
\end{table}

\subsection{Novel Radar View Rendering}
\begin{figure}[t!]
    \centering
    \includegraphics[width=0.99\linewidth]{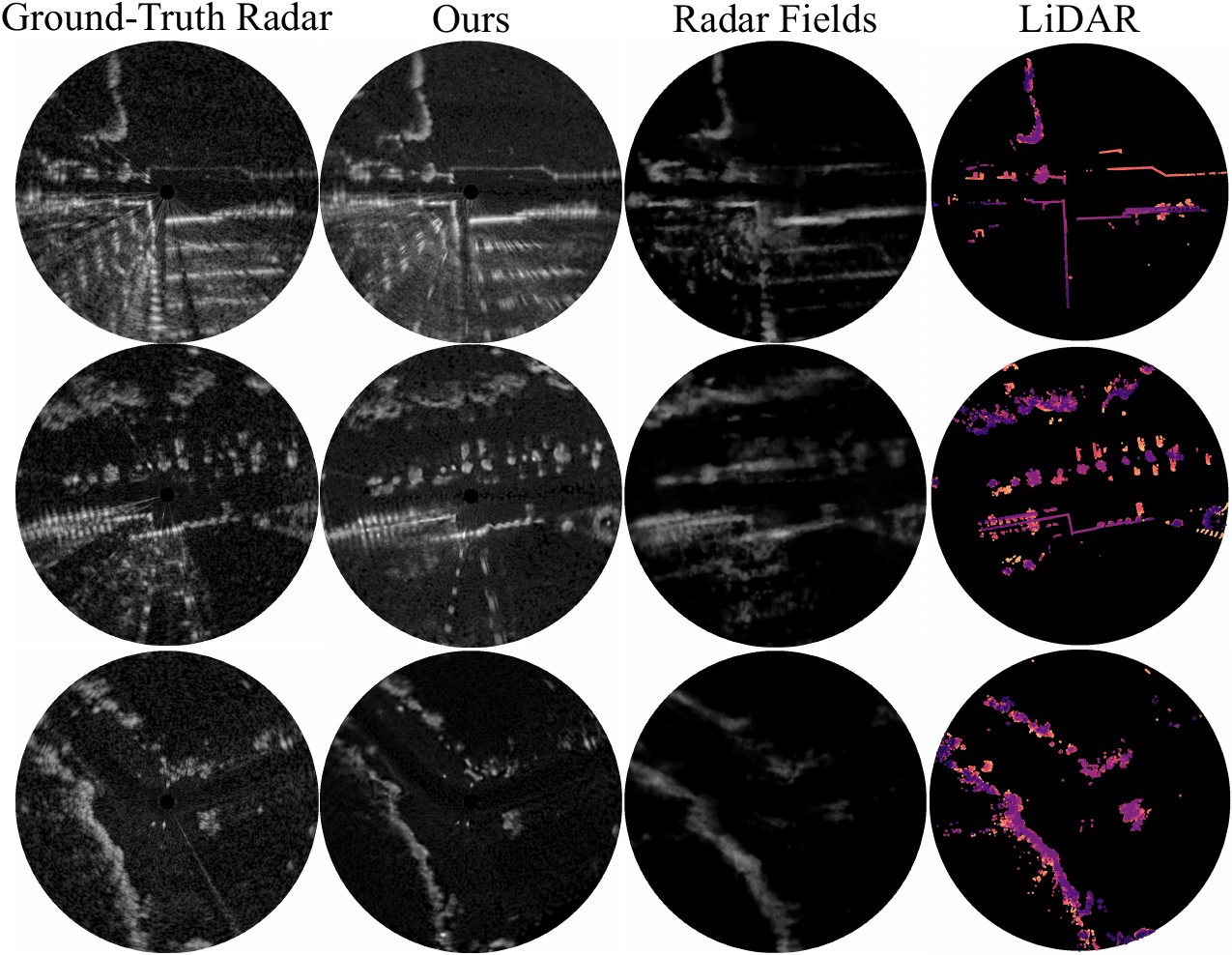}
        \caption{
        %Comparison of radar image rendering from our \algname and Radar Fields, alongside corresponding ground truth radar and LiDAR data. 
        Radar image rendering.
        Our method better synthesizes multipath and noise effects compared to the baseline.}
        \label{fig:img_result}
\end{figure}

\begin{figure}[t!]
    \centering
    \includegraphics[width=0.99\linewidth]{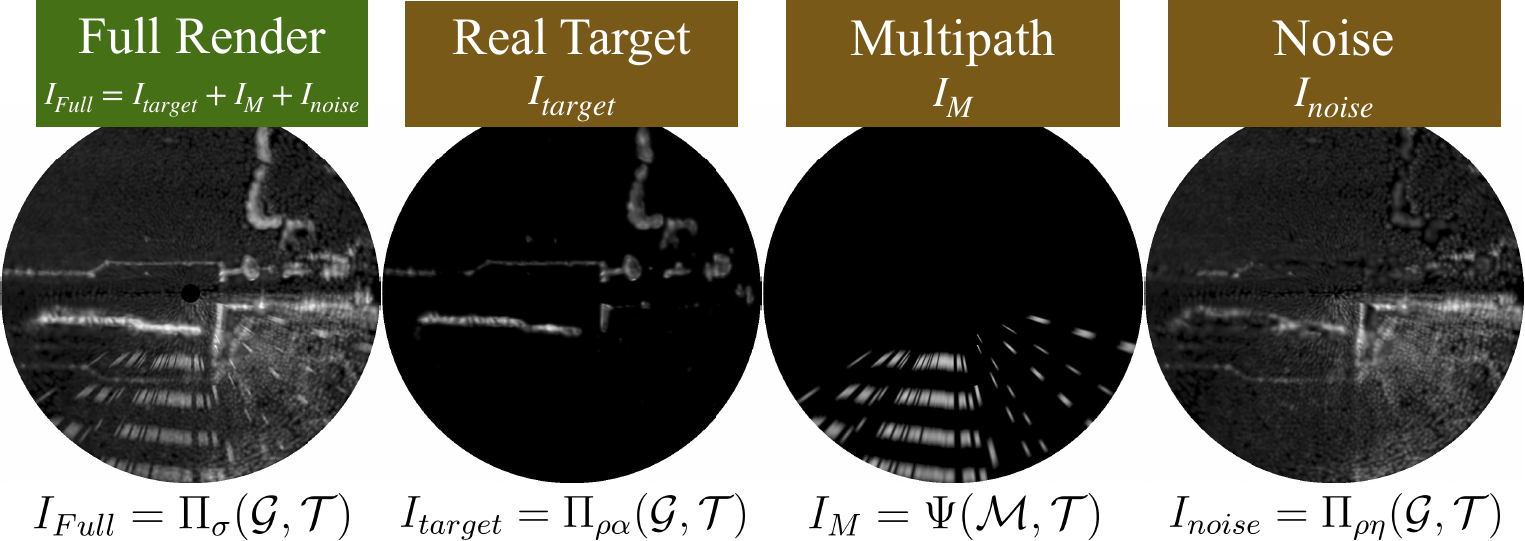}
        \caption{ Radar inverse rendering for radar image decomposition. Our method can disentangle real target render, multipath effect, and other noises (mainly speckle noise and receiver saturation) from radar images.}
        \label{fig:radar_inverse_rendering}
\end{figure}

%\algname provides the capability to render realistic radar images from a novel radar sensor pose.  
We evaluate the synthetic radar image quality with common image rendering metrics, including PSNR, LPIPS, and SSIM. The quantitative evaluation is shown in Table~\ref{tab:vs_radarfields}. With the correct noise modeling and rendering, our proposed method outperforms state-of-the-art, Radar Fields, by $+3.4$ PSNR and achieves more than $2.6\times$ better in SSIM score.
Figure~\ref{fig:img_result} shows the qualitative radar rendering results. Our method accurately renders noise and multipath effects, producing more realistic synthetic radar data. In contrast, Radar Fields fails to model the noise, resulting in noticeable performance degradation. We further demonstrate radar inverse rendering in Figure~\ref{fig:radar_inverse_rendering}. \algname can decompose radar images into different sources.  
The radar image rendering speed reaches 4.5 FPS on an NVIDIA A6000 GPU.

\subsection{Occupancy State Estimation}
While reconstructing noise in the scene, \algname still supports accurate occupancy state estimation by decoupling the occupancy and noise probability from the power return ratio of the Gaussians. To assess the quality of occupancy estimation, we report the RMSE, Relative Chamfer Distance (R-CD), and Accuracy.
%, which is computed as the harmonic mean of precision and recall rate.
%, reflecting the balance between correct positive predictions and overall classification performance. 
Accuracy is computed using a 0.5 m threshold. Ground truth geometry is obtained from the LiDAR point cloud map, which is cropped according to the radar field of view. The point cloud is then projected onto a 2D BEV point cloud for evaluation, following~\cite{rf}. For the Radar Fields occupancy estimation, we follow the original implementation, integrating occupancy along the elevation axis. For \algname, we use the rendered $I_{occ}$ as the final occupancy estimation. A 0.5 probability threshold is then applied to the output of both \algname and Radar Fields to generate the final occupancy map for evaluation.

\begin{figure}[t!]
    \centering
    \includegraphics[width=0.99\linewidth]{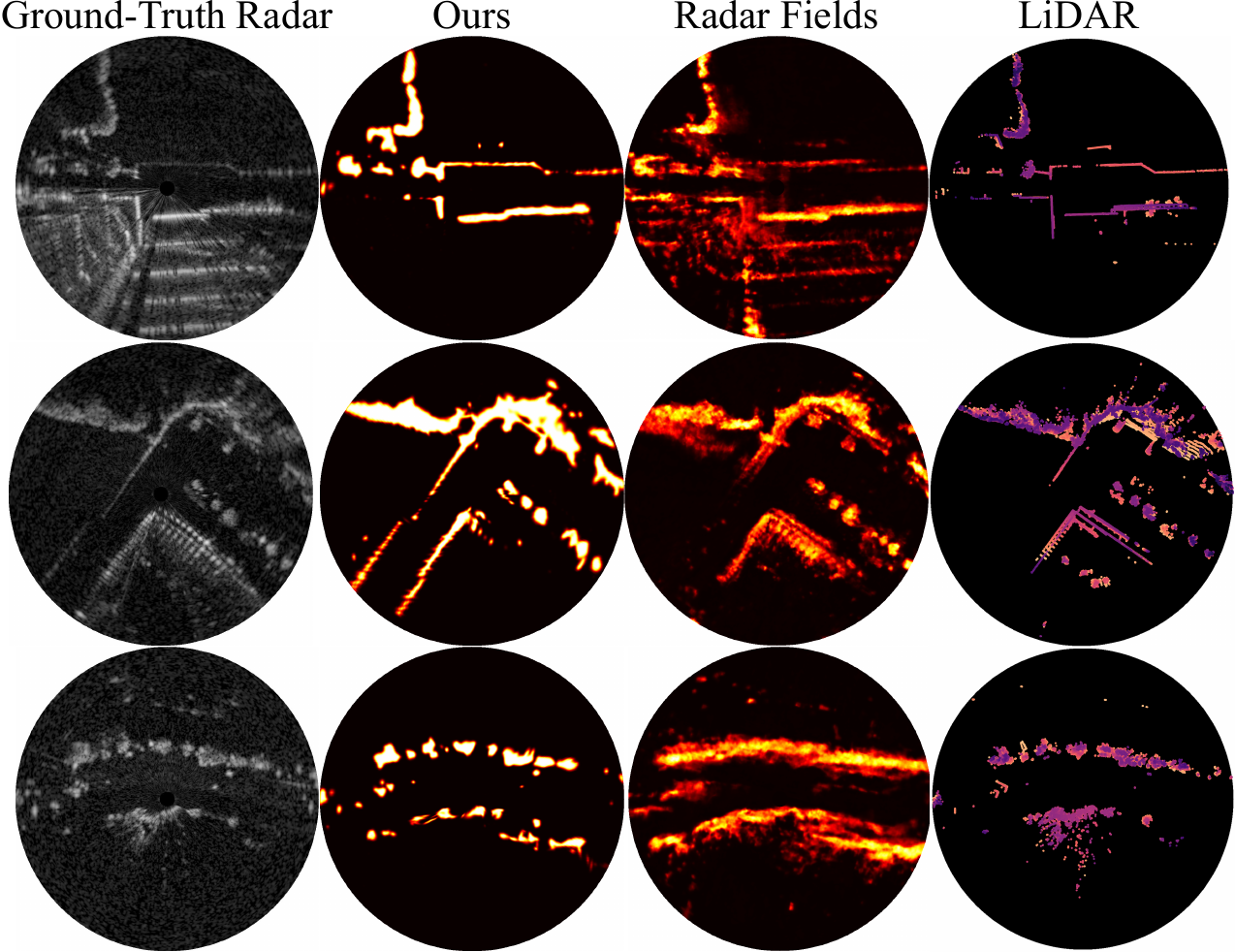}
        \caption{%Comparison of occupancy estimation from our \algname and Radar Fields with corresponding ground truth radar and LiDAR data. 
        Occupancy estimation.
        Our method provides clear and noise-free occupancy estimation compared to the baseline.}
        \label{fig:occ_result}
\end{figure}

\begin{figure*}[t!]
    \centering
    \includegraphics[width=0.99\linewidth]{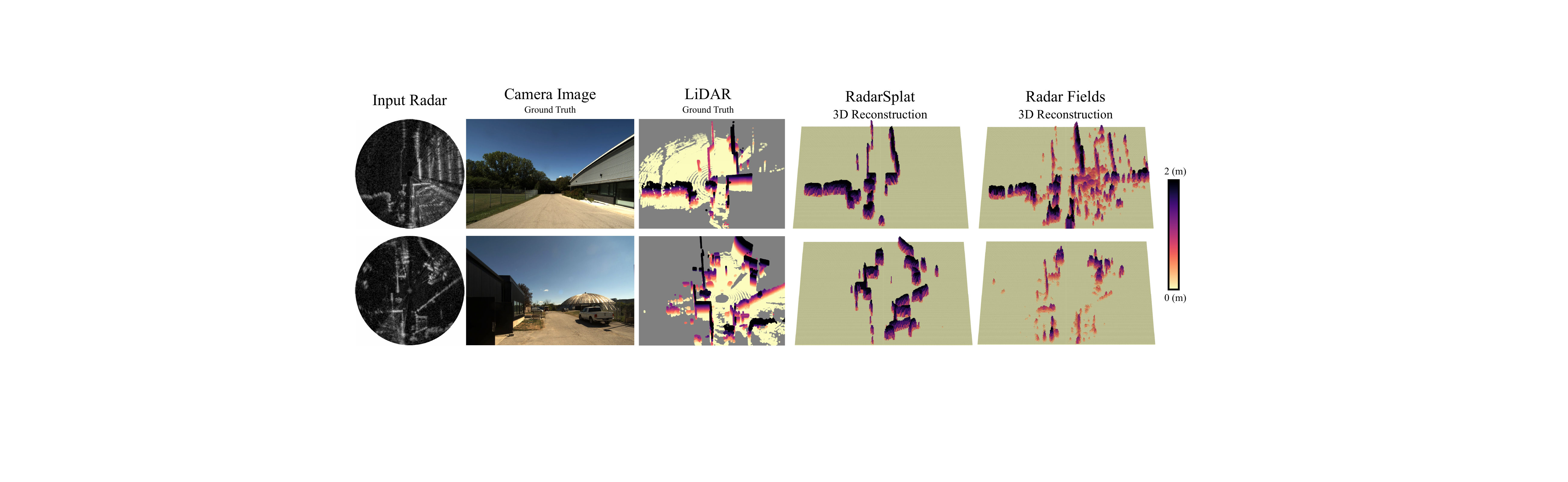}
        \vspace{-0.05in}
        \caption{Comparison of 3D reconstructions from our proposed method \algname and Radar Fields~\cite{rf}. Our method exhibits greater robustness to noise by leveraging multipath modeling and noise probability estimation, which Radar Fields struggles with. }
        \vspace{-0.1in}
        \label{fig:3d_reconstruction}
\end{figure*}

The quantitative evaluation is presented in Table~\ref{tab:vs_radarfields}, alongside the qualitative comparison shown in Figure~\ref{fig:occ_result}. The proposed method outperforms Radar Fields across all metrics, achieving better reconstruction by reducing RMSE by 1.22 m and improving accuracy more than $1.5\times$ compared to Radar Fields. We found that Radar Fields can be significantly affected by severe multipath effects, leading to ghost artifacts behind the wall. Also, it provides blurry occupancy estimations due to saturation and speckle noise. While Radar Fields mitigates most noise using a dynamic threshold during preprocessing and applies Bayesian grid mapping to generate the occupancy grid map, the simple dynamic threshold appears insufficient to effectively handle radar noise in challenging conditions.

In addition, we provide qualitative 3D scene reconstruction results in Figure~\ref{fig:3d_reconstruction}. The results indicate that \algname achieves accurate 3D reconstruction similar to LiDAR, by taking only 2D noisy radar images as input.

\subsection{Ablation Studies}

\begin{table}[b]
\centering
\setlength{\tabcolsep}{4pt}
\renewcommand{\arraystretch}{1.1}
\resizebox{0.8\linewidth}{!}{%
\begin{tabular}{cccccc}
\hline
\multicolumn{2}{c}{Image Synthesis}           & PSNR$\uparrow$ & SSIM$\uparrow$ & LPIPS$\downarrow$ \\ \hline
\multirow{3}{*}{\algname} & w/o Noise Prob. & 23.52          & 0.23           & 0.59              \\
                            & w/o Multipath   & 25.95          & 0.50           & 0.38              \\
                            & Full Method     & \textbf{26.06} & \textbf{0.51}  & \textbf{0.37}    \\ \hline
\end{tabular}
}
\vspace{-0.1in}
\caption{\footnotesize Ablation studies on image synthesis.}
\label{tab:ablation_img}
\end{table}
\begin{table}[b]
\centering
\setlength{\tabcolsep}{4pt}
\renewcommand{\arraystretch}{1.1}
\resizebox{0.9\linewidth}{!}{%
\begin{tabular}{ccccc}
\hline
\multicolumn{2}{c}{Scene Reconstruction}                        & RMSE$\downarrow$ & R-CD$\downarrow$ & Acc.$\uparrow$ \\ \hline
\multirow{2}{*}{Init Occ. Map}           & Radar Fields       & 3.40                 & 0.12                 & 0.90             \\
                                         & Proposed           & 1.81        & 0.04        & 0.90             \\ \hline
\multirow{3}{*}{\algname} & w/o Occ. Map         & 1.86                 & 0.23                 & 0.30             \\
                                         & w/o Spectral Leakage & 2.05                 & 0.05                 & 0.90             \\
                                         & Full Method & \textbf{1.81}        & \textbf{0.04}        & \textbf{0.91}    \\ \hline
\end{tabular}
}
% Init occ. map thres0.15: 1.818	0.042	0.73
% Init occ. map thres0.1: 3.19	0.1016	0.633
% RF occ. map thres0.15: 4.592 0.2001 0.608
\vspace{-0.1in}
\caption{\footnotesize Ablation studies on scene reconstruction.}
\label{tab:ablation_recon}
\end{table}

\begin{figure}[t!]
    \centering
    \includegraphics[width=0.99\linewidth]{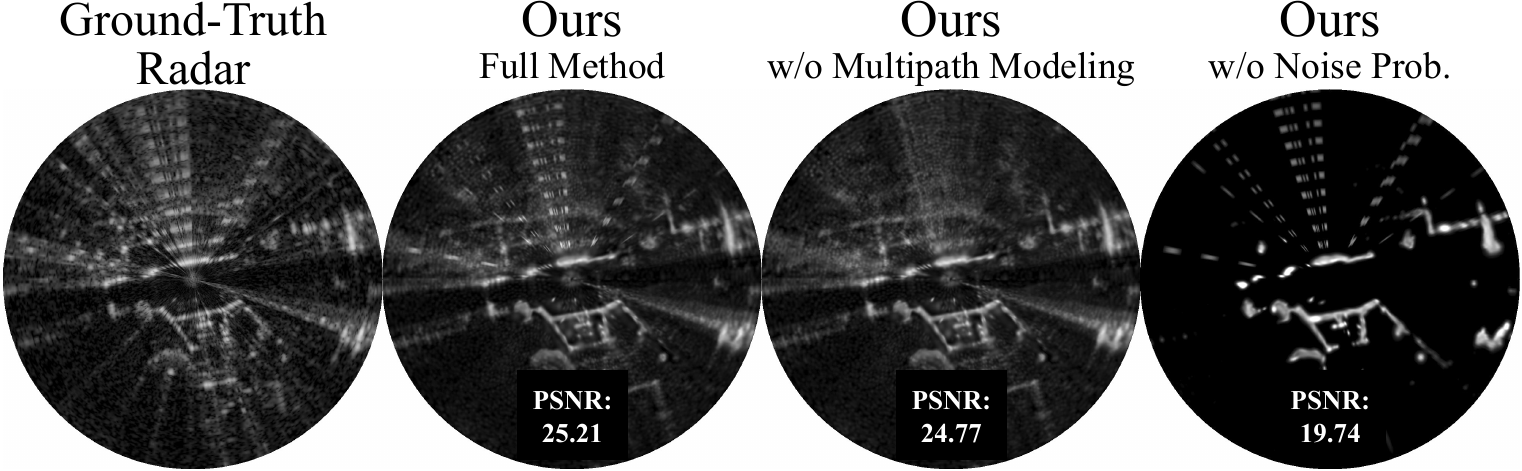}
        \vspace{-0.05in}
        \caption{Ablation studies on image synthesis. \algname fails to model multipath effects when disabling 
 the proposed multipath modeling. 
 %Despite enabling high-order SH for view-dependent reflectivity, the multipath effect is hard to model due to the discontinuity. 
\algname also fails to model other noises when disabling the proposed noise probability.}
        \label{fig:ablations}
\end{figure}

%We also conduct ablation studies to evaluate the effectiveness of the proposed modules. 
Table~\ref{tab:ablation_img} presents the results of our ablation study on image synthesis. We disable multipath modeling (Sec.~\ref{sec:multipath_modeling}) and noise probability (Sec.~\ref{sec:radargs}), which are key features we propose for more realistic rendering. We found that disentangling occupancy and noise probability has a significant impact on radar image synthesis, as speckle noise and receiver saturation frequently occur in radar images. While multipath modeling also improves rendering quality, its quantitative impact is less pronounced compared to noise probability due to the lower frequency of multipath effects. However, in frames with severe multipath interference, it provides substantial qualitative improvements (Figure~\ref{fig:ablations}).

Table~\ref{tab:ablation_recon} presents the results of our ablation study on scene reconstruction. 
In this study, we compare with results with and without spectral leakage modeling (Sec.~\ref{sec:spectral_leakage_modeling}) and occupancy map supervision. we also compare the quality of proposed and existing occupancy map. 
% Our occupancy map provides better initial geometry, demonstrating the efficiency of our denoising.
Our findings indicate that the proposed occupancy map significantly reduces RMSE by nearly $2\times$, as occupancy mapping with a dynamic threshold and a Bayesian grid map alone is insufficient to handle severe radar noise. Also, the occupancy map supervision improves $3\times$ \algname reconstruction accuracy.
%, and using the proposed occupancy map outperform occupancy map used in \cite{rf}. 
On the other hand, spectral leakage modeling also reduce about $0.2$ RMSE. This is expected since the Gaussian variance $\sigma_{w}$ derived from the radar we use is about $0.17$ m.
%In contrast, spectral leakage has a relatively small impact on geometric metrics. This is expected since the Gaussian variance $\sigma_{w}$ derived from the radar we use is only about $0.17$ m. Although the numerical evaluation shows a less pronounced impact, the difference can be substantial in real-world applications that require high reconstruction precision or when using radars with more severe spectral leakage.

\subsection{Adverse Weather and Lighting Conditions}
\begin{figure}[h!]
    \centering
    % \vspace{-0.3in}
    \includegraphics[width=0.99\linewidth]{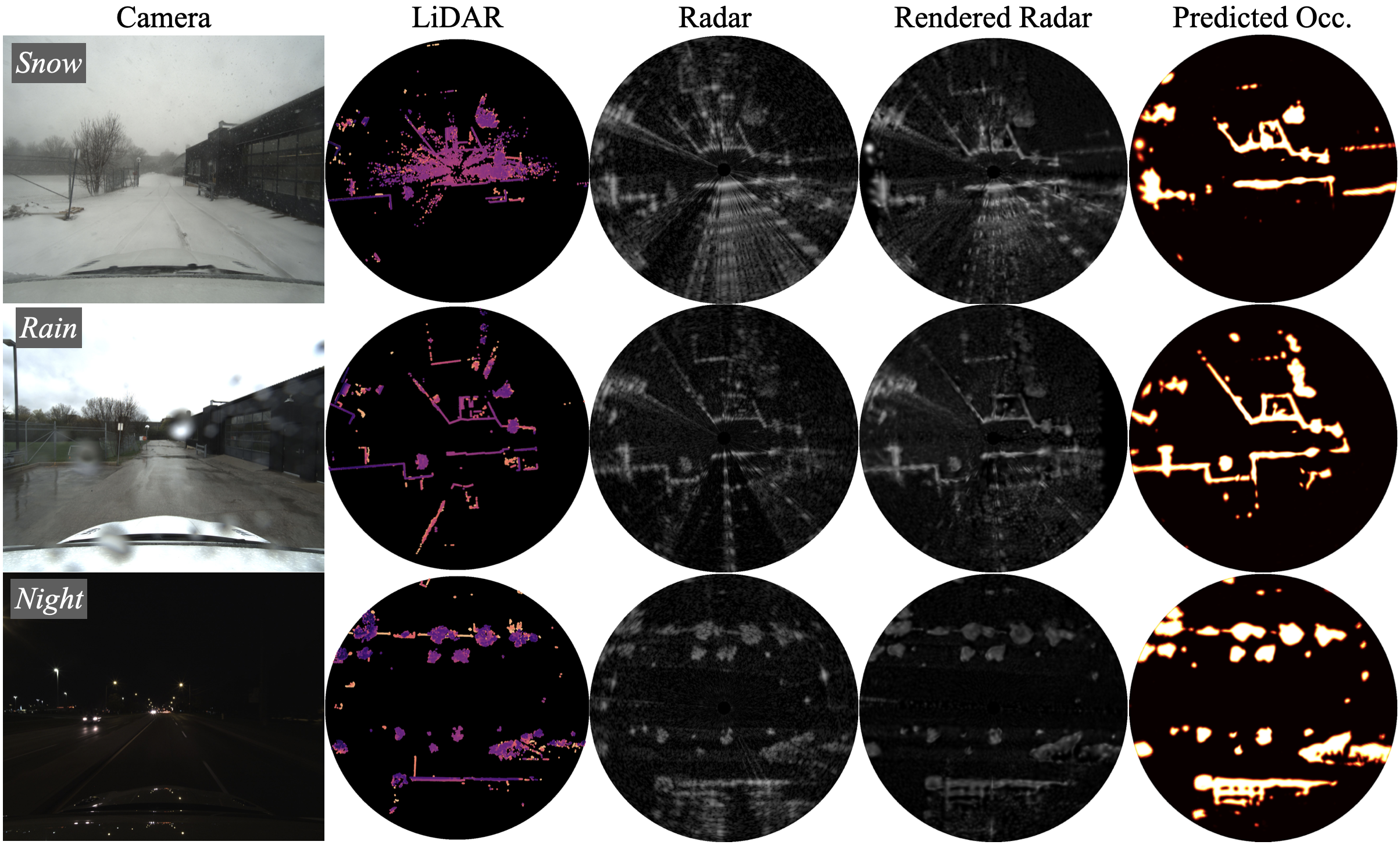}
        % \vspace{-0.1in}
        \caption{Robust radar occupancy estimation in extreme conditions, where either the LiDAR or camera sensor is degraded.}
        \vspace{-0.2in}
        \label{fig:diff_scenes}
\end{figure}

To further demonstrate the robustness of \algname under various weather conditions, Figure~\ref{fig:diff_scenes} illustrates novel view rendering and occupancy maps in snow, rain, and nighttime scenarios. In the snow scene, the LiDAR point cloud exhibits significant artifacts caused by snowfall. In the rain and night scenes, the camera is either blurred due to raindrops or has limited visibility due to low illumination. However, the radar data remains unaffected by these extreme conditions, and the proposed \algname continues to provide robust occupancy map estimations.

\section{Conclusion}
We propose \algname, a Gaussian Splatting method for realistic radar image rendering and accurate occupancy prediction. We take advantage of GS for better rendering quality and present a rendering process that takes radar's unique physical properties into account. Furthermore, we propose a novel noise detection and removal method. In addition, the different types of noise are modeled in a Gaussian primitive and multipath source maps. This enables radar inverse rendering for radar signal decomposition, high-fidelity radar data synthesis, and robust noise-free occupancy prediction. The proposed \algname outperforms the state-of-the-art in both image synthesis and scene reconstruction in both qualitative and quantitative evaluation. We also demonstrate high-quality radar 3D reconstruction with similar performance to LiDAR.

\textbf{Limitations.} The current method only works in static scenes. We plan to extend this work to dynamic scenes, similar to \cite{drivinggs, streetgs}, in the future.

{
    \small
    \bibliographystyle{ieeenat_fullname}
    \bibliography{main}
}

\maketitlesupplementary

In the supplementary material, we first cover the radar sensing primer, radar spectral leakage modeling, and details of the radar used in this study (Sec.~\ref{sec:supp_radar}). Next, we provide additional method details and illustrations for noise detection, denoising, and occupancy grid mapping (Sec.~\ref{sec:supp_noise}). In Sec.~\ref{sec:supp_implementation}, we elaborate on RadarSplat training and provide an overview figure. Sec.~\ref{sec:supp_evaluation} presents evaluation details and additional quantitative and qualitative results. Finally, we discuss the limitations of our current method (Sec. ~\ref{sec:supp_limitation}).

\section{Radar Sensing Primer}
\label{sec:supp_radar}
\subsection{Frequency Modulated Continuous Wave (FMCW) Radar}
\begin{figure}[t!]
    \centering
    \includegraphics[width=0.99\linewidth]{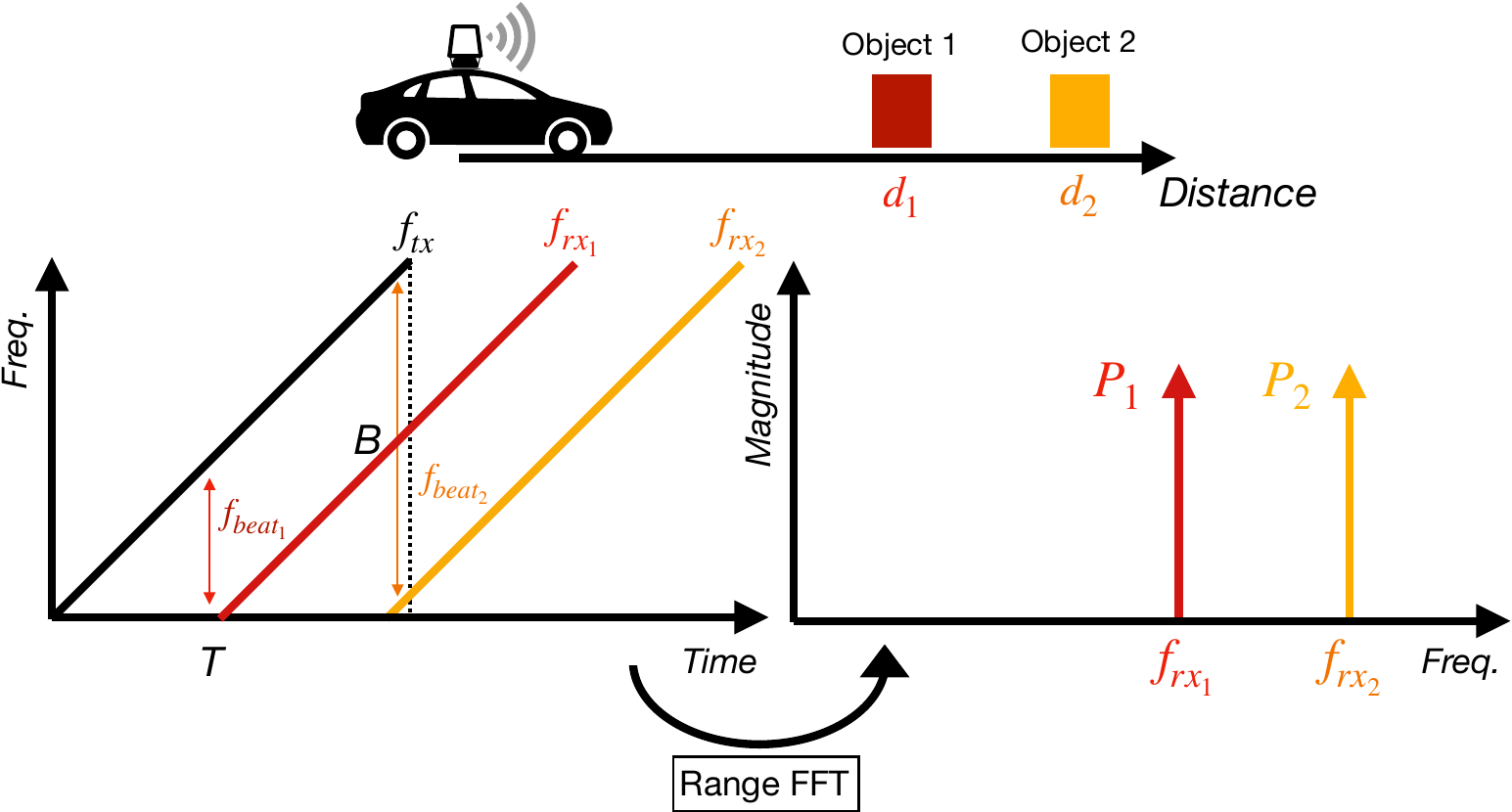}
        \caption{Radar imaging with range fast Fourier transform (FFT).}
        \label{fig:range_fft}
\end{figure}

In FMCW radar systems, the transmitted signal is a linear frequency-modulated chirp. The most common chirp is with the sawtooth pattern. The designed chirp slope is related to the bandwidth $B$ and chirp duration $T$, where the chirp slope is $\frac{B}{T}$. See Figure~\ref{fig:range_fft} for more details~\cite{fmcw_radar}.

Since the radar frequency changes over time, the wave travel time between the target and radar can be measured by the frequency difference between the emitted signal $s_{tx}(t)$ and the returned signal $s_{rx}(t)$, known as beat frequency $f_{beat} = f_{tx}-f_{rx}$, where $f_{tx}$ and $f_{rx}$ are frequencies of transmitted and received waves at time $t$. In practice, the beat frequency can be extracted from the intermediate frequency (IF) signal and a low pass filter (LPF).

Here, we consider a single target example to simplify the explanation. 
% \begin{equation}
%     S(t) = e^{j2\pi(f_0 t+ \frac{B}{2T}t^2)}
% \end{equation}
The IF signal is the complex mixing process between $s_{tx}(t)$ and $s_{rx}(t)$:
\begin{equation}
\begin{split}
S_{IF}(t) = s_{tx}(t) \otimes s_{rx}(t) = \cos(2\pi f_{tx} t) \cdot \cos(2\pi f_{rx} t) = \\ 
\frac{1}{2} \left[ \cos\left[2\pi (f_{tx} - f_{rx})t\right] + \underset{\text{later \space removed \space by \space LPF}}{\underbrace{\cos\left[2\pi (f_{tx} + f_{rx})t\right]}} \right],
\end{split}
\end{equation}
\noindent Next, the frequency-power signal, $F$, can be obtained by a Fast Fourier Transform (FFT), $\mathcal{F}$, also called a range FFT (Figure~\ref{fig:range_fft}). Ideally, the beat frequency $\pm f_{beat}$ should have two impulses:
\begin{equation}
F(f) = \mathcal{F}\{S_{IF}(t)\} = \frac{1}{2}[\delta(f-f_{beat})+\delta(f+f_{beat})]
\label{eq:two_impluses}
\end{equation}
In the end, the range-power signal is obtained by converting frequency $f$ to distance $D$ following:
\begin{equation}
    D = \frac{cf}{2\mu}
\label{eq:freq2dist}
\end{equation}
where $\mu = (B/T)$ is the radar chirp slope.
The radar image we use as input and aim to synthesize consists of multiple range-power signals measured in different azimuth angles with a rotating mechanism.

\subsection{Spectral Leakage Modeling}
\label{sec:spectral_leakage}
\begin{figure}[t!]
    \centering
    \includegraphics[width=0.7\linewidth]{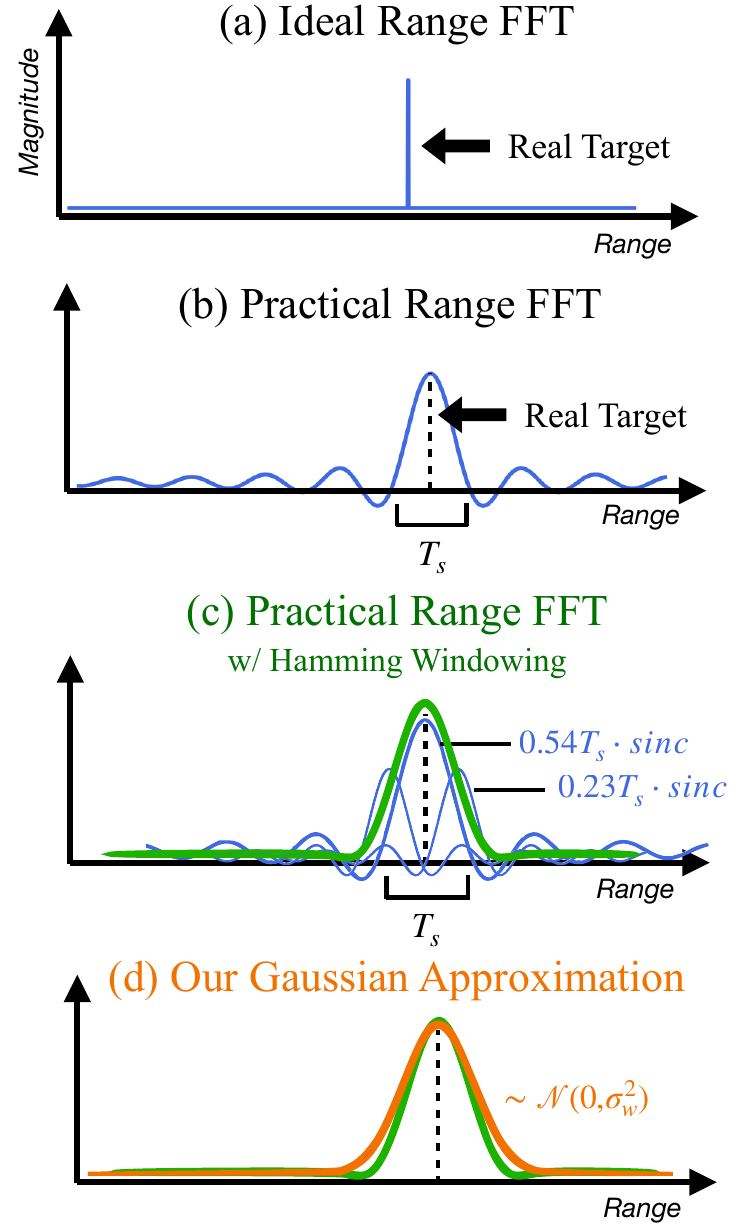}
        \vspace{-0.1in}
        \caption{Modeling spectral leakage in the radar image. (a) Ideal range FFT. (b) Practical range FFT with spectral leakage. (c) The practical range FFT is sharpened using a Hamming window. The final distribution consists of three scaled and shifted sinc functions, as shown.  (d) Proposed Gaussian approximation for Hamming-window-sharpened FFT.}
        \vspace{-0.05in}
        \label{fig:spectral_leakage}
\end{figure}

The spectral leakage is due to finite-time sampling of received signals, causing energy to spread across adjacent frequencies when doing range FFT. 
It transforms the target measurement into a Sinc function (Figure~\ref{fig:spectral_leakage}-b): 
\begin{align}
    \mathcal{F}(f_{IF}(t)) &= \operatorname{sinc} (\omega T_s)
\end{align}
where $\omega=2 \pi f$ and $T_s$ is the sampling duration of radar.

In practice, radar manufacturers apply a windowing technique to obtain sharp frequency cutoff and lower sidelobes. The Navtech radar used in our experiments uses Hamming windowing, which has a range FFT result shown in Figure~\ref{fig:spectral_leakage}-c. The distribution after windowing is: 

\begin{align}
\mathcal{F}(W_{\text{Hamming}}(f(t))) &= T_s \bigg[ 
    0.54 \operatorname{sinc} \left( \frac{\omega T_s}{2\pi} \right) \notag \\
    &\hspace{1.5em} - 0.23 \sum_{k=\pm1} \operatorname{sinc} \left( \frac{\omega T_s + 2k\pi}{2\pi} \right) 
\bigg].
\end{align}

We observe that the radar signal after windowing is similar to a Gaussian distribution. 
Therefore, we use a Gaussian distribution to approximate the blurred effect, as shown in Figure~\ref{fig:spectral_leakage}-d.  
The width of the Sinc function in the frequency domain is:
\begin{equation}
    f_w = \frac{2\pi}{T_s}
\end{equation}
where $T_s$ is the sampling duration of the radar.
The width of the Gaussian in meters $d_w$ is derived with Eq.~\ref{eq:freq2dist}.
We define the variance of the approximated Gaussian as $\sigma_w=(0.5 \times d_w)/3$ so that 99\% of the Gaussian covers the width. Therefore the Gaussian approximation is $\sim \mathcal{N}(0, \sigma_w^2)$.

\begin{figure}[t!]
    \centering
    \includegraphics[width=0.99\linewidth]{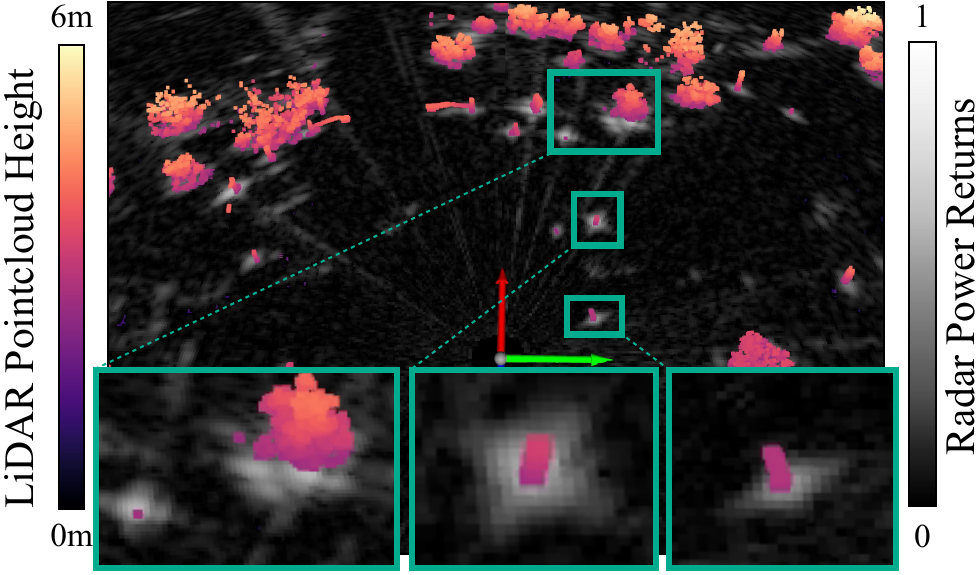}
        \vspace{-0.1in}
        \caption{Illustration of blurred range-power signal due to spectral leakage. The overlapped LiDAR and radar data highlight the effect of radar spectral leakage.}
        % \vspace{-0.05in}
        \label{fig:spectral_leakage_fig}
\end{figure}

The radar used in this paper has $T_s=565 \mu \text{S}$ and $\mu=1.6\times10^{12}$. 
Therefore, we have $d_w\approx1.04~m$ and we set $\sigma_w=0.17~m$ in our experiments. Figure~\ref{fig:spectral_leakage_fig} illustrates the spectral leakage effect by visualizing overlapped radar image and LiDAR point cloud.

\subsection{Scanning Radar Details}
\begin{figure}[t!]
    \centering
    \includegraphics[width=0.99\linewidth]{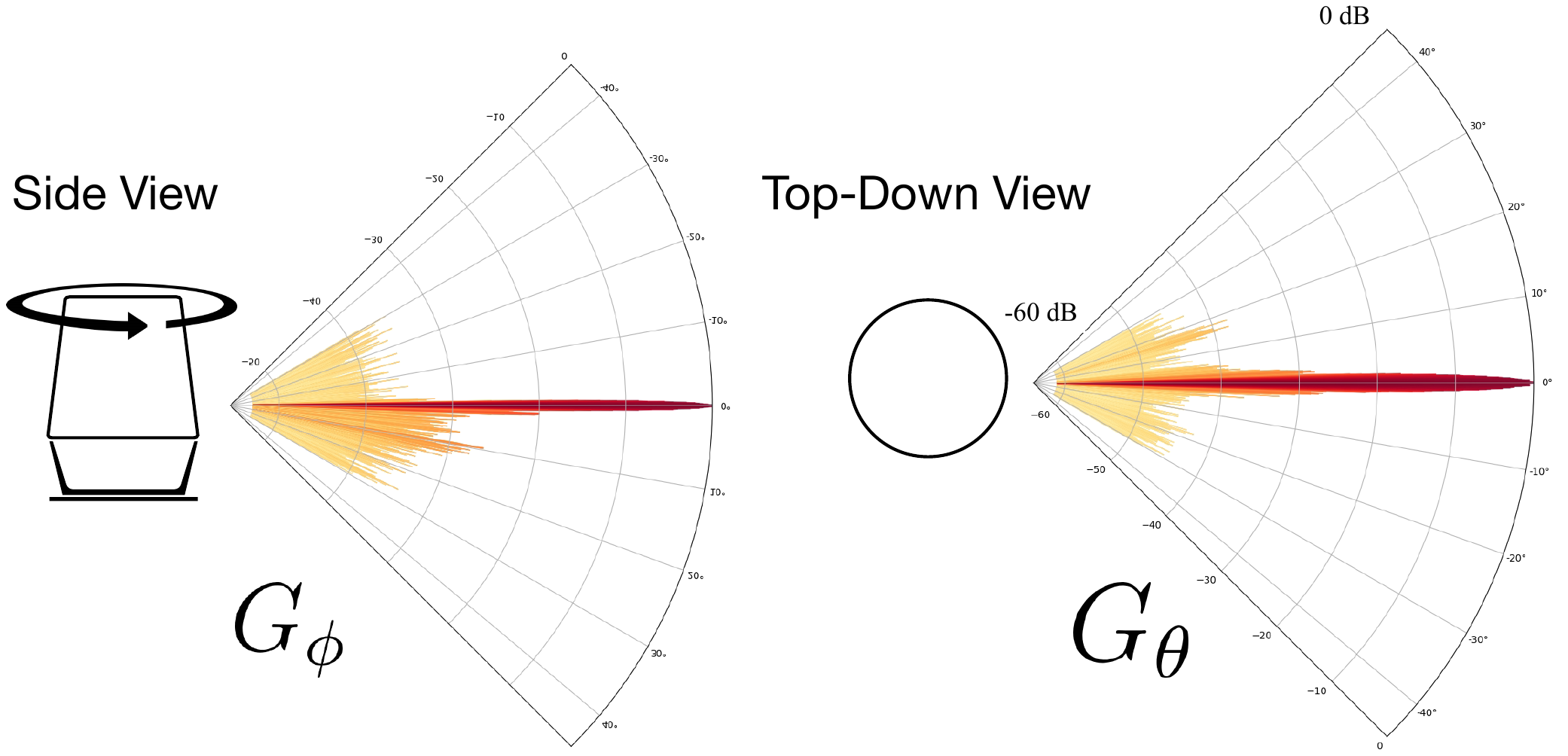}
        \caption{Visualize azimuth and elevation radar antenna gain.}
        \label{fig:antenna_gain}
\end{figure}

\begin{figure}[t!]
    \centering
    \includegraphics[width=0.9\linewidth]{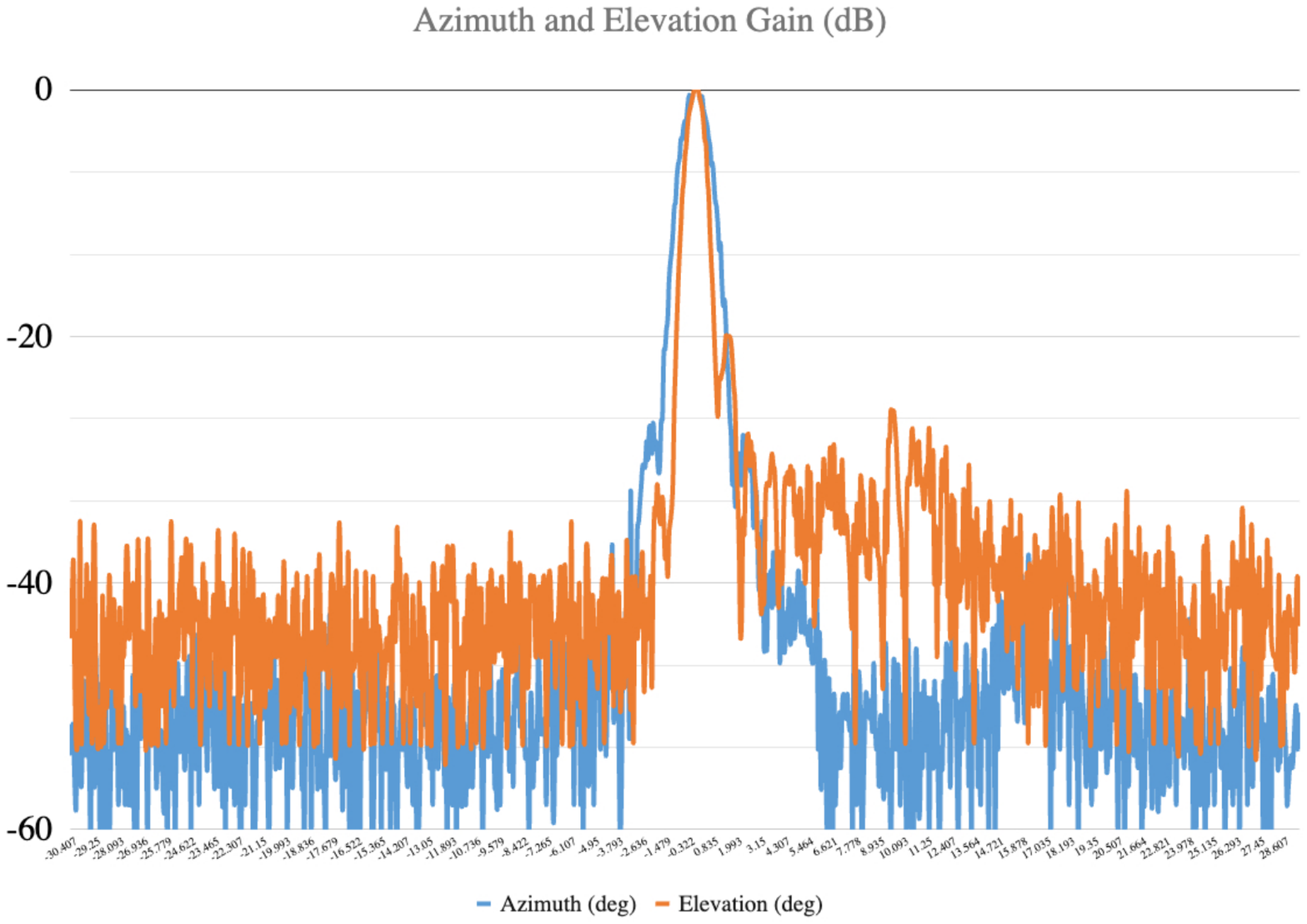}
        \caption{Antenna gain plot.}
        \label{fig:antenna_gain_plot}
\end{figure}

The radar used in this paper is Navtech CIR304-H from the Boreas Dataset~\cite{boreas}. The radar is operated at 4Hz scanning rate. The sequence we used has 0.0596 m range resolution and 0.9 horizontal resolution. The beam spread is $1.8^\circ$ between -3 dB attenuation points horizontally and vertically. 
Additionally, the vertical antenna gain is designed with a cosec squared fill-in beam pattern, which enables a wider elevation field of view (FOV) to up to 40 degrees below the sensor plane.
The radar antenna gain provided by the radar manufacturer is shown in Figure~\ref{fig:antenna_gain}. The detailed numbers in dB is shown in Figure~\ref{fig:antenna_gain_plot}.

\section{Noise Detection, Denoising, and Occupancy Mapping}
\label{sec:supp_noise}

\subsection{Multipath and Saturation Noise Detection}
\begin{figure*}[t]
    \centering
    \includegraphics[width=0.9\linewidth]{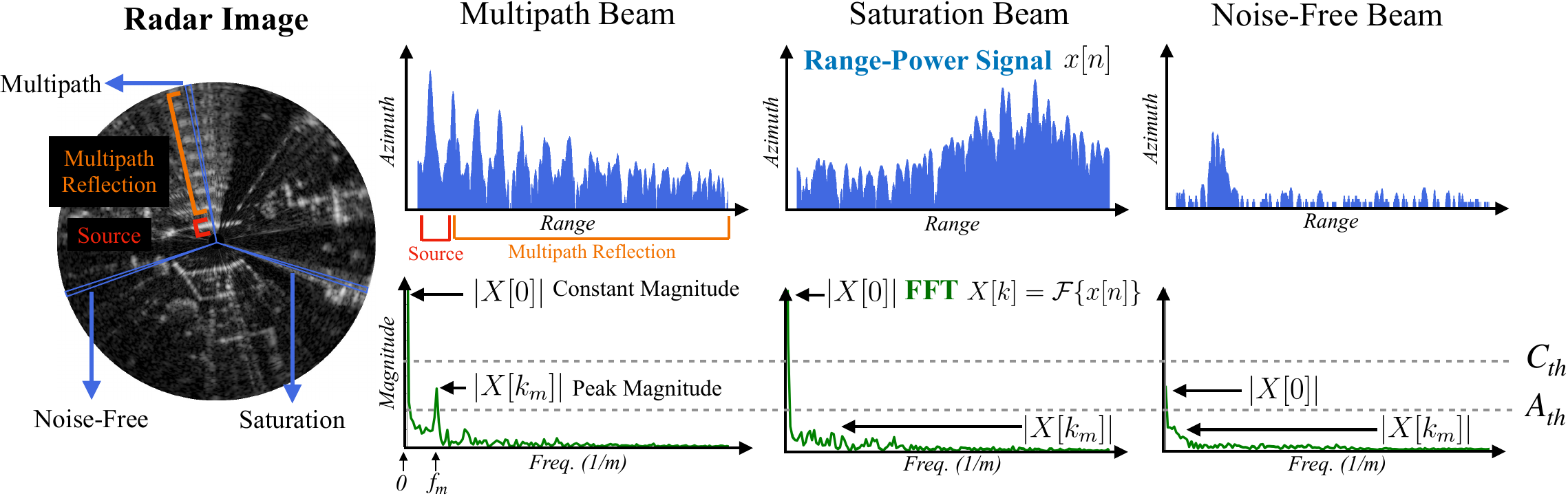}
        \caption{Illustration of the range-power signal of multipath, receiver saturation, and noise-free azimuth beam and their FFT. Also the constant and peak magnitude is shown along with the constant ratio threshold $C_{th}$ for noise detection and with the peak magnitude threshold $A_{th}$ for multipath detection.}
        \label{fig:FFT_full}
\end{figure*}

Figure~\ref{fig:FFT_full} shows the range-power signal and FFT of multipath, receiver saturation, and noise-free azimuth beams. The saturation beam is detected when constant ratio $C>C_{th}$. The multipath is detected when $C>C_{th}$ and $|X[k_m]|>A_{th}$.

\subsection{Denoising}
We reconstruct an initial estimated occupancy map to guide training, following~\cite{rf}. Instead of relying solely on a dynamic threshold, our noise detection enables more robust noise removal, leading to improved initial occupancy maps for training. 
We propose a denoising algorithm that removes noise across detected noisy azimuth angles, $\theta_{noise} \in \Theta_{sat} \cup \Theta_{multi}$.
We first apply Gaussian smoothing on the range-power signal to reduce the impact of high-frequency noise.
\begin{equation}
     G_{\sigma_s}(n) = \frac{1}{\sqrt{2\pi\sigma_s^2}} e^{-\frac{n^2}{2\sigma_s^2}}
\end{equation}
\begin{equation}
     x_{smooth}[n] = (x[n]*G_{\sigma_s}),
\end{equation}
where the $\sigma_s$ is the variance of Gaussian, which we set to 5 bins in our implementation. Next, a masking region selection algorithm is applied to $x_{smooth}[n]$. We find the range index with the maximum magnitude and search the decay region to generate a noise-free mask region $(n_{s}, n_{e})$. Every bin $n$ outside $(n_{s}, n_{e})$ region is set to zero. The pseudo-code is shown in Algorithm.~\ref{algo:denoise}. The denoised image is then used to construct the initial occupancy map.

\begin{algorithm}[t]
\caption{Denoising algorithm with Decay Regions in Radar Range-Power Data}
\label{alg:find_decay_region}
\begin{algorithmic}[1]
\Require $P(n)$ (1D array of radar power values), $\sigma_s$ (Gaussian smoothing parameter)
\Ensure $n_{\max}$ (Index of maximum power), $\mathcal{D} = (n_s, n_e)$ (Decay region start and end indices)
\State $P_{\text{smooth}}(n) \gets \text{GaussianFilter1D}(P(n), \sigma_s)$
\State $n_{\max} \gets \arg\max(P_{\text{smooth}}(n))$
\State $n_s \gets n_{\max}$
\While{$n_s > 0$ \textbf{and} $P_{\text{smooth}}(n_s - 1) \leq P_{\text{smooth}}(n_s)$}
    \State $n_s \gets n_s - 1$
\EndWhile
\State $n_e \gets n_{\max}$
\While{$n_e < \text{length}(P_{\text{smooth}}(n)) - 1$ \textbf{and} $P_{\text{smooth}}(n_e + 1) \leq P_{\text{smooth}}(n_e)$}
    \State $n_e \gets n_e + 1$
\EndWhile
\State \Return $(n_s, n_e)$
\end{algorithmic}
\label{algo:denoise}
\end{algorithm}

\subsection{Occupancy Mapping}
To the handle occlusion issue, we use a $W$ frames window to reconstruct an occupancy map for each frame. We set $W=10$ in our experiment. We compute the power mean of each grid to identify the free and occupied space. A power threshold $p_{th}=0.15$ is chosen to obtain a binary grid map for training supervision.

\section{RadarSplat Implementation Details}
\label{sec:supp_implementation}

\subsection{Input Format}
We set the maximum range of the input radar image to 50 m, with an azimuth resolution of 0.9° and a range resolution of 0.0596 m, resulting in an input image size of (400, 839). To ensure accurate error computation, we mask out the closest 2.5 m of radar data, as these measurements primarily originate from the ego-vehicle.

\subsection{RadarSplat}
\begin{figure*}[t!]
    \centering
    \includegraphics[width=0.99\linewidth]{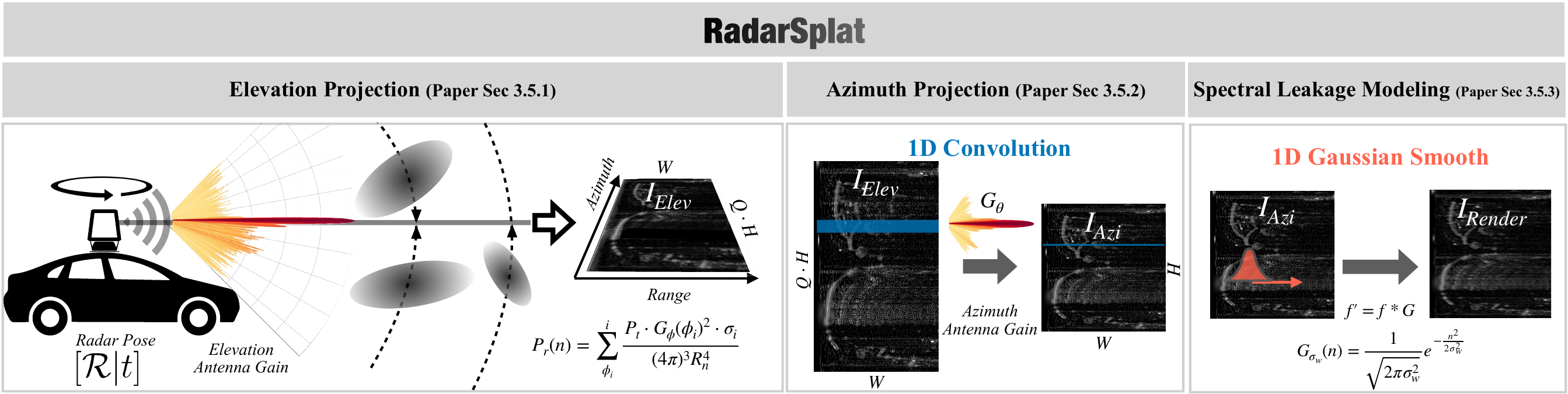}
        \caption{\algname rendering. We first project Gaussians to 2D using the elevation antenna gain, then apply the azimuth antenna gain via a 1D convolution along the azimuth axis. Radar spectral leakage is modeled using 1D Gaussian smoothing along the range axis.}
        \label{fig:radar_rendering}
\end{figure*}

The \algname rendering pipeline, illustrated in Figure~\ref{fig:radar_rendering}, consists of three key stages: elevation projection, azimuth projection, and spectral leakage modeling. In practice, we set $Q=10$ in the elevation projection step, which results in the rendered elevation-projected image $I_{Elev}$ having a size of $(4000, 839)$. The azimuth projection is implemented using a 1D convolution along the azimuth axis with a stride size of Q and circular padding. As a result, $I_{Azi}$ has a size of (400, 839). In the end, the spectral leakage modeling is applied with Gaussian variance $\sigma_w=0.17~m$ derived from Sec.~\ref{sec:spectral_leakage}.

For Gaussian Splatting, we initialize $2\times10^4$ Gaussians with occupancy probability $\alpha=0.1$ and noise probability $\eta=0.1$. The initial Gaussian size is set to 0.5~m with random initialization. The spherical harmonics level is set to 10. %Although DART~\cite{dart} reports using 25-level spherical harmonics, we only observe incremental improvements at such high level when evaluating novel view rendering.

\subsection{Training Configuration}
We set $\lambda_1=0.8, \lambda_2=0.2, \lambda_3=5, \lambda_4=10^2, \lambda_5=10^2$. \algname is initialized with $2\times10^4$ Gaussians of size $s=0.5~m$ and trained for $3000$ iterations. For multipath modeling, we set $C_{th}=0.21, A_{th}=0.3, C'_{th}=0.2, r_{th}=0.5m, \text{ and } \theta_{th}=10^\circ$. Gaussian rendering is set to $Q=10$.

\subsection{Cartesian-to-spherical Gaussian Conversion}
The conversion is as follows:
\begin{align}
\mu_{spherical}=
\begin{bmatrix}
    r \\
    \theta \\
    \phi
\end{bmatrix}
=
\begin{bmatrix}
    \sqrt{x^2 + y^2 + z^2} \\
    \arctan2(y, x) \\
    \arcsin\left(\frac{z}{r}\right)
\end{bmatrix}
\end{align}
\begin{equation}
    \Sigma_{\text{spherical}} = J \Sigma J^T
\end{equation}
where $J$ is the Jacobian of Cartesian-to-spherical space conversion.

\begin{align}
J =
\begin{bmatrix}
\frac{x}{r} & \frac{y}{r} & \frac{z}{r} \\
-\frac{y}{x^2 + y^2} & \frac{x}{x^2 + y^2} & 0 \\
-\frac{xz}{r^2\sqrt{r^2 - z^2}} & -\frac{yz}{r^2\sqrt{r^2 - z^2}} & \frac{\sqrt{r^2 - z^2}}{r^2}
\end{bmatrix}
\end{align}

\section{Evaluation Details and Extra Evaluation}
\label{sec:supp_evaluation}

% \subsection{Image Synthesis Evaluation}
% In image synthesis evaluation, we use widely adopted metrics such as PSNR, SSIM, and LPIPS. While these metrics are reasonable for evaluating full images, radar images pose unique challenges. Most radar image pixels represent low-power returns and empty space, making conventional metrics less effective. Additionally, radar noise, such as multipath reflection, is a critical factor in generating realistic radar images. However, since multipath beams occur less frequently than multipath-free beams, their impact on standard metrics is limited.

% To address this limitation and provide a more radar-specific evaluation, we propose the Multipath PSNR (MP PSNR) metric, which exclusively evaluates radar beams detected as multipath, ensuring a more accurate assessment of radar image synthesis quality. The evaluation results using the MP PSNR metric are presented in Table~\ref{tab:vs_radarfields_full}.

% With the MP PSNR metric, we can clearly distinguish the performance differences between Radar Fields~\cite{rf} and the proposed method. Radar Fields exhibits a significant performance drop on multipath beams due to the absence of multipath modeling. In contrast, the proposed method demonstrates only a minor 0.11 PSNR drop on these challenging multipath beams, highlighting its robustness in handling multipath effects.

\subsection{Scene Reconstruction Evaluation}
We construct a LiDAR pointcloud map to obtain ground-truth geometry for evaluation. Similar to building an occupancy map, we use a W-frame window to reconstruct a local map for each frame, preventing occluded objects from being included in the ground-truth map. To further solve the occlusion problem and radar invisible objects in the scene (mostly tree leaves and sticks), we adopt a small radar power threshold of $0.1$ to remove all the LiDAR points having corresponding radar measurements below the threshold. Also, the $1.8^\circ$ elevation angle is applied when obtaining LiDAR ground truth for evaluation. Here are details about our scene reconstruction metrics:

\textbf{Relative Chamfer Distance (R-CD)}. The Relative Chamfer Distance normalizes the Chamfer Distance by the maximum pairwise distance between the predicted point cloud, $P$, and the ground truth point cloud, $Q$. The Chamfer Distance is defined as:
\begin{align}
    CD(P, Q) = \frac{1}{|P|} \sum_{p \in P} \min_{q \in Q} \| p - q \|_2^2 + \notag \\
    &\hspace{-10.0em}\frac{1}{|Q|} \sum_{q \in Q} \min_{p \in P} \| q - p \|_2^2
\end{align}
The Relative Chamfer Distance is defined as:
\begin{equation}
R\text{-}CD(P, Q) = \frac{CD(P, Q)}{\max\limits_{q_i, q_j \in Q_{}} \| q_i - q_j \|_2^2}
\end{equation}

\textbf{Accuracy}.
Accuracy is computed as the ratio of correctly matched points (both in precision and recall sense) over the total number of points in both clouds.
\begin{equation}
    \footnotesize 
    \text{Accuracy} = \frac{| \{ p \in P \mid d(p, Q) < \tau \} | + | \{ q \in Q \mid d(q, P) < \tau \} |}{|P| + |Q|}
\end{equation}
where $\tau$ is the distance threshold and 
\begin{equation}
    d(p, Q) = \min\limits_{q \in Q} \| p - q \|_2 
\end{equation}
is the nearest neighbor distance from each predicted point to the ground truth. We set $\tau=0.5$ in practice.
The accuracy can be divided into precision and recall:

\textbf{Precision}.
Precision measures the fraction of reconstructed points that are within \( \tau \) of the ground truth:
\begin{equation}
    \text{Precision} = \frac{| \{ p \in P \mid d(p, Q) < \tau \} |}{|P|}
\end{equation}

\textbf{Recall}.
Recall measures the fraction of ground truth points that have a corresponding reconstructed point within \( \tau \):
\begin{equation}
    \text{Recall} = \frac{| \{ q \in Q \mid d(q, P) < \tau \} |}{|Q|}
\end{equation}

%The evaluation results with precision and recall metrics are presented in Table~\ref{tab:vs_radarfields_full}. 
Here we provide a more detailed evaluation with precision and recall, as shown in Table~\ref{tab:vs_radarfields_full}. The results show that we have significant improvement in both precision and recall compared to baseline. However, we observed that precision is relatively lower than recall, indicating that our occupancy estimation has more false positives than false negatives. We hypothesize that the primary reason for this is the reconstruction of certain structures that are occluded in the LiDAR point cloud but visible in the radar image.

\begin{table}[h!]
\centering
\setlength{\tabcolsep}{4pt}
\renewcommand{\arraystretch}{1.3}

\resizebox{0.9\linewidth}{!}{%
\begin{tabular}{clccccc}
\hline
\multirow{2}{*}{Method} &  & \multicolumn{5}{c}{Scene Reconstruction}                                                                        \\ \cline{3-7} 
                        &  & RMSE$\downarrow$ & R-CD$\downarrow$ & Acc.$\uparrow$ & \textbf{Precision$\uparrow$} & \textbf{Recall$\uparrow$} \\ \cline{1-1} \cline{3-7} 
Radar Fields            &  & 3.03             & 0.29             & 0.59           & 0.46                         & 0.61                      \\
Ours                    &  & \textbf{1.81}    & \textbf{0.04}    & \textbf{0.91}  & \textbf{0.71}                & \textbf{0.94}             \\ \hline
\end{tabular}
}
\caption{Scene reconstruction evaluation with precision and recall metrics.}
\label{tab:vs_radarfields_full}
\end{table}

\subsection{Scene-Separated Evaluation}
In Table \ref{tab:scenes}, we show \algname's results in different weather and lighting conditions.
The consistent results across diverse weather and lighting conditions also show robustness of our method. 

\begin{table}[!h]
\centering
\setlength{\tabcolsep}{6pt}
\renewcommand{\arraystretch}{1.2}

\resizebox{1.0\linewidth}{!}{%
\begin{tabular}{lcccclccc}
\hline
\multirow{2}{*}{Scenes} &  & \multicolumn{3}{c}{Image Synthesis}                 &  & \multicolumn{3}{c}{Scene Reconstruction}                                  \\ \cline{3-5} \cline{7-9} 
                        &  & PSNR$\uparrow$ & SSIM$\uparrow$ & LPIPS$\downarrow$ &  & RMSE$\downarrow$ & R-CD.$\downarrow$ & \multicolumn{1}{l}{Acc.$\uparrow$} \\ \cline{1-1} \cline{3-5} \cline{7-9}
\hline
Sunny                                    &                         & 25.81          & 0.50           & 0.37              & \textbf{} & 1.78             & 0.04              & 0.90                               \\
Snowy                                    &                         & 25.79          & 0.52           & 0.36              & \textbf{} & --               & --                & --                                 \\
Rainy                                    &                         & 26.59          & 0.49           & 0.39              & \textbf{} & 1.63             & 0.03              & 0.93                               \\
Night                                    &                         & 26.69          & 0.51           & 0.37              &           & 2.12             & 0.06              & 0.92                               \\ \hline
\end{tabular}
}
\vspace{-0.1in}
\caption{ \algname scene-separated evaluation}
\label{tab:scenes}
\end{table}

\subsection{Ablation Studies for Multipath Modeling on Scenes With and Without Multipath Reflections}
Table~\ref{tab:ablation_urban_natural} show separate ablations on urban scenes with many multipath effects and natural scenes without multipath effects to quantify the contribution of multipath modeling.

\begin{table}[h]
\centering
\setlength{\tabcolsep}{4pt}
\renewcommand{\arraystretch}{1.3}
\resizebox{1.0\linewidth}{!}{%
\begin{tabular}{lcccclccc}
\hline
\multicolumn{2}{c}{\multirow{2}{*}{Image Synthesis}} & \multicolumn{3}{c}{Urban Scene}                     & \multicolumn{1}{c}{} & \multicolumn{3}{c}{Natural Scene}                   \\ \cline{3-5} \cline{7-9} 
\multicolumn{2}{c}{}                                 & PSNR$\uparrow$ & SSIM$\uparrow$ & LPIPS$\downarrow$ &                      & PSNR$\uparrow$ & SSIM$\uparrow$ & LPIPS$\downarrow$ \\ \hline
\multirow{2}{*}{RadarSplat}      & w/o Multipath     & 25.25          & 0.49           & 0.37              &                      & \textbf{27.27} & \textbf{0.52}  & \textbf{0.37}     \\
                                 & Full Method       & \textbf{26.06} & \textbf{0.51}  & \textbf{0.37}     &                      & \textbf{27.27} & \textbf{0.52}  & \textbf{0.37}     \\ \hline
\end{tabular}
}
\caption{ In urban scenes, where multipath reflections are prominent, the modeling improves results. In contrast, in natural scenes with only trees surrounding the area, where multipath effects are minimal, the multipath modeling has negligible impact.}
\label{tab:ablation_urban_natural}
\end{table}

\subsection{Ablation Studies on Occupancy Maps}
We validate the impact of the proposed denoised occupancy map in proposed \algname and Radar Fields. Table~\ref{tab:ablation_hybrid} shows both Radar Fields and our method benefit from the proposed occupancy map.

\begin{table}[!h]
\vspace{-0.1in}
\centering
\setlength{\tabcolsep}{4pt}
\renewcommand{\arraystretch}{1.3}
\resizebox{0.9\linewidth}{!}{%
\begin{tabular}{ccccc}
\hline
\multicolumn{2}{c}{Scene Reconstruction}                 & RMSE$\downarrow$ & R-CD$\downarrow$ & Acc.$\uparrow$ \\ \hline
\multirow{2}{*}{Radar Fields} & w/ RF Occ. Map & 3.03             & 0.29             & 0.59           \\
                              & w/ Proposed Occ. Map     & \underline{2.68}             & \underline{0.11}             & \underline{0.72}           \\ \hline
\multirow{2}{*}{RadarSplat}   & w/ RF Occ. Map & 1.83             & 0.05             & 0.90           \\
                              & w/ Proposed Occ. Map     & \underline{\textbf{1.81}}    & \underline{\textbf{0.04}}
                              & \underline{\textbf{0.91}}  \\ \hline
\end{tabular}
}
\vspace{-0.1in}
\caption{ Comparing effect of ours and Radar Fields' occupancy map.}
\vspace{-0.15in}
\label{tab:ablation_hybrid}
% \vspace{-0.2in}
\end{table}

\subsection{Ablation Studies on Initialized Gaussians}
Table~\ref{tab:ablation_num_gaussian} and \ref{tab:ablation_size_gaussian} show performance improves with increasing number and size of initialized Gaussians, saturating at the chosen 20k Gaussians and 0.5 m size.

\begin{table}[h!]
\centering
\setlength{\tabcolsep}{4pt}
\renewcommand{\arraystretch}{1.1}
\resizebox{0.7\linewidth}{!}{%
\begin{tabular}{ccccl}
\hline
Gaussians Num & 5k          & 10k         & 20k & 30k \\ \hline
PSNR $\uparrow$  & 25.31         & 25.79         & 26.06 & 26.05 \\
Acc. $\uparrow$  & 0.88 & 0.90 & 0.91  & 0.91  \\ \hline
\end{tabular}
}
\caption{Gaussian number ablation.}
\label{tab:ablation_num_gaussian}
\end{table}

\begin{table}[h!]
\centering
\setlength{\tabcolsep}{4pt}
\renewcommand{\arraystretch}{1.1}
\resizebox{0.7\linewidth}{!}{%
\begin{tabular}{ccccl}
\hline
Gaussians Size & 0.1   & 0.3   & 0.5   & 0.7   \\ \hline
PSNR $\uparrow$   & 21.73 & 25.14 & 26.06 & 26.20 \\
Acc. $\uparrow$   & 0.65  & 0.90  & 0.91  & 0.91  \\ \hline
\end{tabular}

}
\caption{Gaussian size ablation.}
\label{tab:ablation_size_gaussian}
\end{table}

% \begin{table}[!h]
% \centering
% \begin{minipage}{0.5\linewidth}
% \centering
% \setlength{\tabcolsep}{3pt}
% \renewcommand{\arraystretch}{1.2}
%   \resizebox{\linewidth}{!}{
%     \input{rebuttal_tables/ablation_num_gaussian}
%   }
%   \caption{ \small Gaussian number ablation. } %\scriptsize
%   \label{retuttal_tab:ablation_num_gaussian}
% \end{minipage}%
% \hfill
% \begin{minipage}{0.5\linewidth}
% \centering
% \setlength{\tabcolsep}{3pt}
% \renewcommand{\arraystretch}{1.1}
%   \resizebox{\linewidth}{!}{
%     \input{rebuttal_tables/ablation_size_gaussian}
%   }
%   \caption{ \small Gaussian size ablation.}
%   \label{retuttal_tab:ablation_size_gaussian}
% \end{minipage}
% \end{table}

\subsection{Additional Results}
\begin{figure*}[!t]
    \centering
    \includegraphics[width=0.9\linewidth]{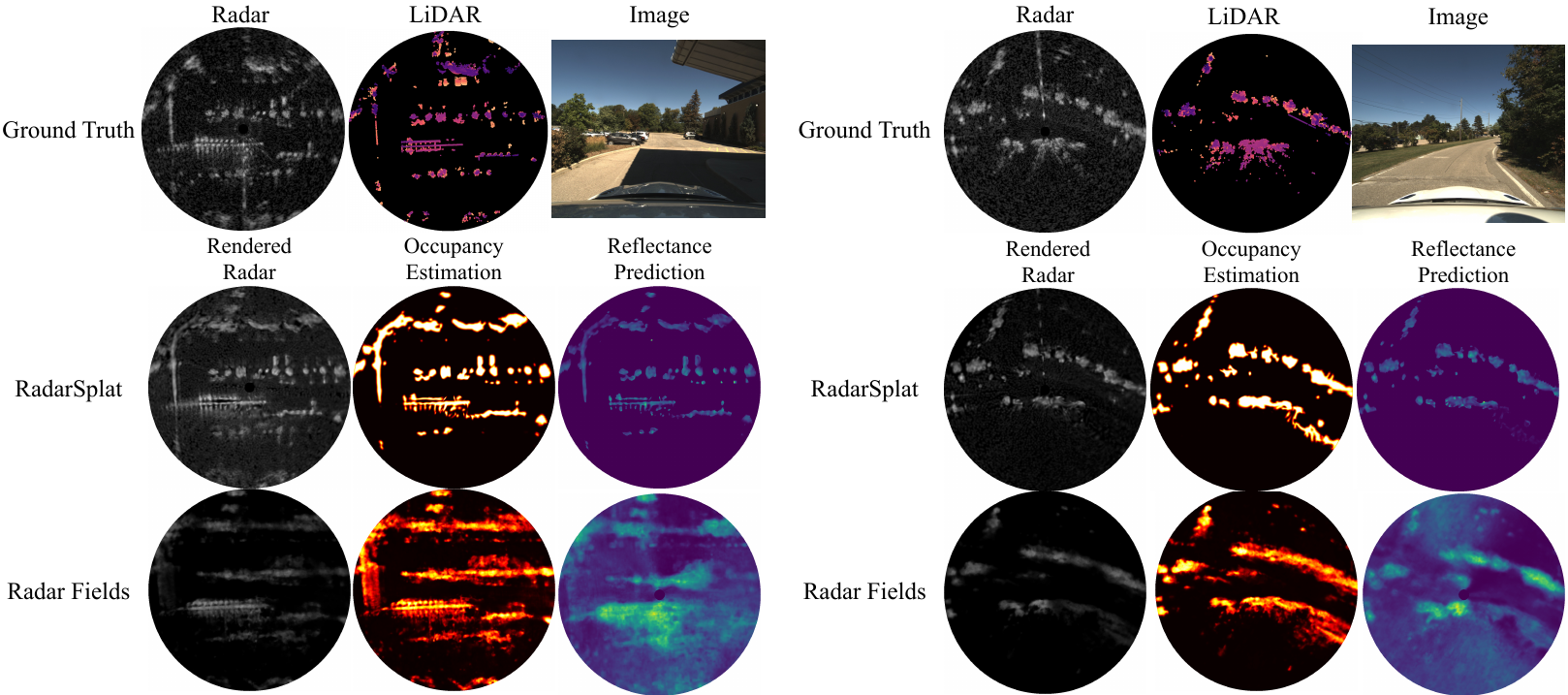}
        \caption{Additional results on Boreas Dataset.}
        \label{fig:supp_results}
\end{figure*}

In Figure~\ref{fig:supp_results}, we present additional results, including the ground-truth camera view and the rendered reflectance. RadarSplat achieves superior image synthesis and occupancy estimation compared to the baseline~\cite{rf}. For reflectance rendering, we take advantage of explicit Gaussian representation to segment out Gaussian that has low occupancy probability and high noise probability, resulting in a clearer object reflectance map. In addition, the videos of radar 3D reconstruction compared with ground-truth LiDAR are provided in the supplementary materials zip file.

\begin{figure*}[!t]
    \centering
    \includegraphics[width=0.9\linewidth]{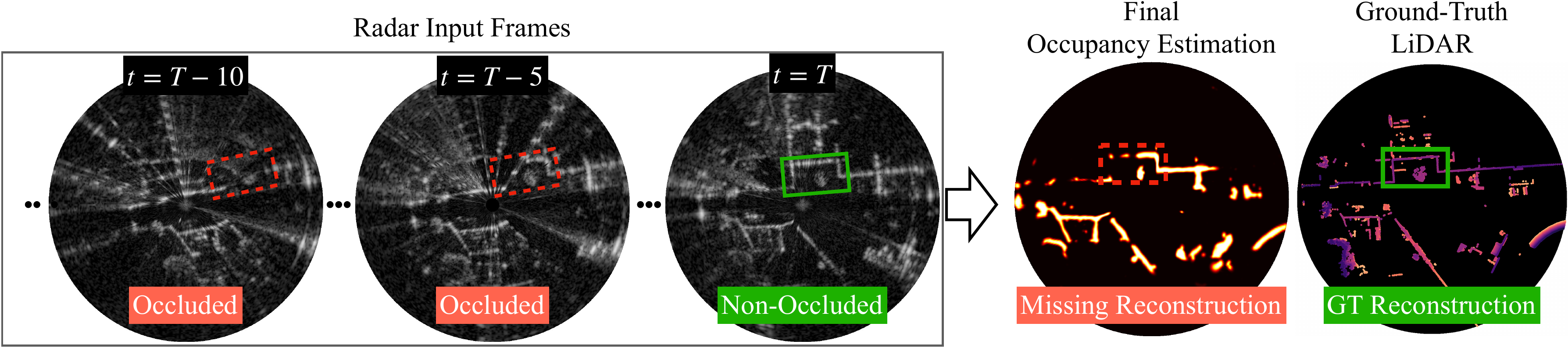}
        \caption{Wrong occupancy estimation caused by the occlusion problem.}
        \label{fig:occlusion_problem}
\end{figure*}

\section{Limitations}
\label{sec:supp_limitation}
\textbf{Occlusion Problem}. Although radar provides bird-eye-view (BEV) power images with radar waves penetrating and bouncing off to see through occluded objects, occlusion can still happen in the radar image if the objects have high reflectivity. Figure~\ref{fig:occlusion_problem} illustrates that a region behind the corner is occluded in most training views, resulting in wrong occupancy estimation. %However, radar occlusion modeling is challenging and required to consider wave penetrating and bounce-off modeling.

% \textbf{Motion Distortion}.
% The scanning radar operates at a low scanning rate of 4 Hz. Similar to LiDAR, radar scans become distorted when the radar is moving too fast because beams at different azimuth angles are captured at different timestamps. However, in our current rendering pipeline, we render all radar beams as if they were captured at the same timestamp. This approximation works well in low-speed scenarios but introduces increasing distortion as speed increases.

\textbf{Dynamic Objects}.
Similar to \cite{rf}, our method does not consider dynamic object modeling. However, the Gaussian scene graph approach proposed in \cite{drivinggs, streetgs} can be incorporated to model moving objects separately as individual Gaussian splats. These dynamic splats can then be combined with static Gaussian splats to construct a complete dynamic scene representation.

In the future, we plan to overcome these limitations by integrating occlusion modeling and the Gaussian scene graph into our method.

% {
%     \small
%     \bibliographystyle{ieeenat_fullname}
%     \bibliography{main}
% }

% \end{document}

\end{document}